\documentclass{article} % For LaTeX2e
\usepackage{iclr2023_arxiv,times}

\usepackage{amsmath,amsfonts,bm}

\def\eqref#1{equation~\ref{#1}}
\def\1{\bm{1}}

\DeclareMathAlphabet{\mathsfit}{\encodingdefault}{\sfdefault}{m}{sl}
\SetMathAlphabet{\mathsfit}{bold}{\encodingdefault}{\sfdefault}{bx}{n}

\usepackage{hyperref}
\usepackage{url}
\usepackage{bm}
\usepackage{algorithm}
\usepackage{algorithmicx}
\usepackage{xspace}
\usepackage{xcolor}
\usepackage{subcaption}
\usepackage{graphicx}
\usepackage{booktabs}
\usepackage{pifont}
\usepackage[export]{adjustbox}

\makeatletter
\DeclareRobustCommand\onedot{\futurelet\@let@token\@onedot}
\def\@onedot{\ifx\@let@token.\else.\null\fi\xspace}
 
\def\eg{\emph{e.g}\onedot} 
\def\ie{\emph{i.e}\onedot}

\def\wrt{w.r.t\onedot}

\makeatother

\newcommand{\cmark}{\ding{51}}%
\newcommand{\xmark}{\ding{55}}%

\makeatletter
    \def\tagform@#1{\maketag@@@{\normalsize(#1)\@@italiccorr}}
\makeatother

\title{Gaussian-Bernoulli RBMs Without Tears}

\author{Renjie Liao\thanks{Work done partially as a visiting faculty researcher at Google Brain.}~~$^{1}$, Simon Kornblith$^{2}$, Mengye Ren$^{3}$, David J.\ Fleet$^{2,4,5}$, Geoffrey Hinton$^{2,4,5}$ \\
University of British Columbia$^1$, Google Research, Brain Team$^2$, \\ 
New York University$^3$, University of Toronto$^4$, Vector Institute$^5$ \\
\texttt{rjliao@ece.ubc.ca, mengye@cs.nyu.edu}\\
\texttt{\{skornblith, davidfleet, geoffhinton\}@google.com}
}

\iclrfinalcopy % Uncomment for camera-ready version, but NOT for submission.
\begin{document}

\maketitle

\begin{abstract}
We revisit the challenging problem of training Gaussian-Bernoulli restricted Boltzmann machines (GRBMs), introducing two innovations. We propose a novel Gibbs-Langevin sampling algorithm that outperforms existing methods like Gibbs sampling.
We propose a modified contrastive divergence (CD) algorithm so that one can generate images with GRBMs starting from noise.
This enables direct comparison of GRBMs with deep generative models, improving evaluation protocols in the RBM literature. 
Moreover, we show that modified CD and gradient clipping are enough to robustly train GRBMs with large learning rates, thus removing the necessity of various tricks in the literature.
Experiments on Gaussian Mixtures, MNIST, FashionMNIST, and CelebA show GRBMs can generate good samples, despite their single-hidden-layer architecture. 
Our code is released at: \url{https://github.com/lrjconan/GRBM}
\end{abstract}

\section{Introduction}\label{sec:intro}

% Restricted Boltzmann machines (RBMs) \citep{smolensky1986information,freund1991unsupervised,hinton2002training} are energy-based generative models with stochastic binary visible units and binary hidden units.
% It is a variant of Boltzmann machines \citep{ackley1985learning} with a bipartite graphical structure, thus enabling efficient probabilistic inference.
% And RBMs can be stacked sequentially to build deep belief networks (DBNs) \citep{hinton2006reducing,bengio2006greedy,hinton06fast} which are widely used deep learning architectures.
% Nevertheless, the binary-data assumption underlying the original RBMs limits their applicability to continuous input data, \eg, images and audios. 
% Gaussian-Bernoulli RBMs (GRBMs) 
% % or GBRBMs), \aka, Gaussian-binary restricted Boltzmann machines, 
% \citep{welling2004exponential,hinton2006reducing} are a natural extension of RBMs, again with a bipartite graphical model, but with continuous visible units and binary hidden units.

Restricted Boltzmann machines (RBMs) \citep{smolensky1986information,freund1991unsupervised,hinton2002training} are  energy-based generative models with stochastic binary units.
A variant of Boltzmann machines \citep{ackley1985learning}, they have a bipartite graphical structure that enables efficient probabilistic inference, and they can be stacked to form deep belief networks (DBNs) \citep{hinton2006reducing,bengio2006greedy,hinton06fast}.
Gaussian-Bernoulli RBMs (GRBMs) \citep{welling2004exponential,hinton2006reducing} extend RBMs to model continuous data by replacing the binary visible units of the RBM with Gaussian random variables.

% Despite their attractive properties, it is widely known that learning GRBMs is quite challenging, and various  modifications have been proposed to mitigate these challenges
GRBMs remain challenging to learn, however, despite many proposed modifications to the model or training algorithm.
% \citep{lee2007sparse,krizhevsky2009learning,ranzato2010factored,cho2011improved,hinton2012practical,cho2013gaussian,melchior2017gaussian,upadhya2021learning}.
For instance, \citet{lee2007sparse} add a regularization term to encourage sparsely activated binary hidden units.
\citet{krizhevsky2009learning} attribute the difficulties in learning to high-frequency noise present in natural images.
Factorized high-order terms were introduced in \citep{ranzato2010modeling,ranzato2010factored} to allow GRBMs to explicitly learn the covariance structure among pixels.
\citet{nair2010rectified} suggest that binary hidden units are problematic,
and proposed model variants with real-valued hidden units. 
\citet{cho2011improved,cho2013gaussian} advocate the use of parallel tempering sampling \citep{earl2005parallel}, adaptive learning rate, and enhanced gradient \citep{cho2011enhanced} to improve GRBM learning.
\citet{melchior2017gaussian} conclude that difficulties in GRBM training are due to training algorithms rather than the model itself; they advocate the use of gradient clipping, specialized weight initialization, and contrastive divergence (CD) \citep{hinton2002training} rather than persistent CD \citep{tieleman2008training}.
\citet{upadhya2021learning} propose a stochastic difference of convex functions programming (S-DCP) algorithm to replace CD in training GRBMs. 

An important motivation for seeking to improve GRBM learning is so that a GRBM can be used 
% as a front-end
to convert real-valued data to stochastic binary data. This would make it easy for researchers to explore novel ways of implementing stochastic binary Boltzmann machines to model real-valued data. 
To that end, we propose improved GRBM learning methods for image data.  Specifically,
\begin{itemize}
    \itemsep 0.05cm
    
    \item We propose a hybrid Gibbs-Langevin sampling algorithm that outperforms predominant use of Gibbs sampling. To the best of our knowledge this is the first use of Langevin sampling for GRBM training (with or without Metropolis adjustment).
    
    \item We propose a modified CD algorithm so that one can generate images with learned GRBMs starting from Gaussian noise. 
    This enables a fair and direct comparison of GRBMs with  deep generative models, something beyond the reach of existing GRBM learning methods.
  
    \item We show that the modified CD  with gradient clipping is sufficient to train GRBMs, thus removing the need for heuristics that have been crucial for existing approaches. 
    
    \item We empirically show that GRBMs can generate good samples on Gaussian Mixtures, MNIST, FashionMNIST, and CelebA, despite they have a single hidden layer.
\end{itemize}

\section{Related Work}\label{sec:relate_work}

% Learning GRBMs is known to be hard and various analysis and improvements have been proposed.
% We summarize the main difficulties below and differentiate existing methods with ours accordingly.

\textbf{Learning the variances }
Learning the variance of visible units in GRBMs is necessary 
for generating sharp and realistic images.  But small variances tend to cause the energy function and its gradient to have large values, thus making the stochastic gradient estimates returned by CD numerically unstable.
Most existing methods fix the variance (\eg, to one) to avoid this issue.
\citet{krizhevsky2009learning,cho2011improved} consider learning the variance using a smaller learning rate than for other parameters, obtaining much better reconstruction, thus supporting the importance of learning variances.
However, many of the learned filters are still noisy and point-like.
\citet{melchior2017gaussian} learn a shared variance across all visible units, yielding improved performance,
especially with large numbers of hidden units.
In this work, we learn one variance parameter per visible unit and achieve much lower learned variances than existing methods, \eg, approximately $1e^{-5}$ on MNIST.

% Since the distribution of visible units conditioned on hidden units is Gaussian in GRBMs, one would hope to learn small variances so that the visible units are not too noisy, \eg, images look sharp.
% However, small variances would make the energy function and its gradient numerically large, thus making the stochastic gradient estimates returned by CD numerically unstable.
% Most existing methods fix the variance (\eg, to one) to avoid this issue.
% \citet{krizhevsky2009learning,cho2011improved} consider learning the variance using a smaller learning rate than for other parameters, obtaining much better reconstruction, thus supporting the importance of learning variances.
% However, many of the learned filters are still noisy and point-like.
% \citet{melchior2017gaussian} learn a shared variance across all visible units, yielding improved performance,
% especially with large numbers of hidden units.
% In this work, we learn one variance parameter per visible unit and achieve much lower learned variances than existing methods, \eg, approximately $1e^{-5}$ on MNIST.

\textbf{Stochastic gradient estimation and learning rate }
Due to the intractable log partition function of GRBMs, one often estimates the gradients of the log likelihood \wrt parameters via Monte Carlo.
Gibbs sampling is predominant in CD learning due to its simplicity, but it mixes slowly in practice.  This yields noisy gradient estimates which often cause training instabilities and prohibits using large learning rates.
\citet{cho2011improved} explore parallel tempering with adaptive learning rates to obtain better reconstruction.
\citet{cho2013gaussian} propose enhanced gradients that are invariant to bit-flipping in hidden units.
\citet{melchior2017gaussian} show that gradient clipping and special weight initialization support robust CD learning with large learning rates.
We advocate Langevin MC to improve gradients, and validate that gradient clipping does enable training with large learning rates.

\textbf{Model capacity }
\citet{theis2011all} empirically show that GRBMs are outperformed even by simple mixture models in estimating likelihoods for image data.
\citet{wang2012analysis,melchior2017gaussian} demonstrate that GRBMs can be expressed as either a product of experts or a constrained Gaussian mixture in the visible domain, hinting that GRBMs need more hidden units than the true number of components to fit additive mixture densities well.
\citet{krause2013approximation,gu2022approximation} provide theoretical guarantees on GRBMs for universal approximation of mixtures and smooth densities.
Although this shows that GRBMs are expressive, they do not lead directly to practical GRBM learning algorithms.

\textbf{Model Evaluation }
Like many deep generative models, evaluating GRBMs is difficult, as the log likelihood is intractable.
To date, GRBMs have been evaluated by visually inspecting reconstructed images, filters and hidden activation (\ie, features), and sampled images during CD training.
Quantitative metrics include reconstruction errors, and error rates of post-hoc trained classifiers on learned features.
However, these metrics do not necessarily indicate if GRBMs are good generative models \cite{melchior2017gaussian}.
Unlike existing work, we sample from learned GRBMs, starting from Gaussian noise, enabling direct comparisons with other generative models, qualitatively (visually inspecting samples) and quantitatively (\eg, Frechet Inception distance (FID) \citep{heusel2017gans}.

% \section{Model}
% \label{sec:model}

% \subsection{Gaussian-Bernoulli Restricted Boltzmann Machines (GRBMs)}
% \label{subsec:GRBM}

\section{Gaussian-Bernoulli Restricted Boltzmann Machines}
\label{sec:GRBM}

A Gaussian-Bernoulli Restricted Boltzmann Machine (GRBM) \citep{welling2004exponential, krizhevsky2009learning, cho2011improved, melchior2017gaussian} is a  Markov Random Field (MRF) with continuous stochastic visible units and binary stochastic hidden units.
% GRBMs extend Restricted Boltzmann Machines (RBMs) to model the density of continuous data rather than binary data.
Denoting $N$ visible units as $\mathbf{v} \in \mathbb{R}^{N}$ and $M$ hidden units as $\mathbf{h} \in \{0, 1\}^{M}$, the energy function associated with a GRBM  is defined to be
\begin{align}\label{eq:energy_GRBM}
    E_{\theta}(\mathbf{v}, \mathbf{h}) = \frac{1}{2} \left( \frac{\mathbf{v} - \bm{\mu} }{ \bm{\sigma} } \right)^{\top}
    \left( \frac{\mathbf{v} - \bm{\mu} }{ \bm{\sigma} } \right) - \left( \frac{\mathbf{v}}{\bm{\sigma}^2} \right)^{\top} W \mathbf{h} - \mathbf{b}^{\top} \mathbf{h} \, 
%      E_{\theta}(\mathbf{v}, \mathbf{h}) = \frac{1}{2} \left( \mathbf{v} - \bm{\mu} \right)^{\top} C^{-1}
%   \left(\mathbf{v} - \bm{\mu} \right) - \mathbf{v}^{\top} C^{-1}W \mathbf{h} - \mathbf{b}^{\top} \mathbf{h}
   \, ,
\end{align}
with 
% $C = \mathrm{diag}( {\bm{\sigma}})^2  $ is the covariance matrix, 
weight matrix $W \in \mathbb{R}^{N \times M}$,  bias $\bm{b} \in \mathbb{R}^{M}$, mean $\bm{\mu} \in \mathbb{R}^{N}$, and variance $\bm{\sigma}^2 \in \mathbb{R}_{+}^{N}$,
where, unless stated otherwise, 
$\frac{\mathbf{x}}{\mathbf{y}}$ denotes element-wise division between  vectors $\mathbf{x}$ and $\mathbf{y}$, as is convention in the GRBM literature.
% \footnote{I.e.,  $\frac{\mathbf{x}}{\mathbf{y}}$ is equivalent to $\diag({\mathbf{y}})^{-1} \mathbf{x}$, where $\diag(\mathbf{y})$ denotes the diagonal matrix formed from a vector $\mathbf{y}$.}
We denote the set of learnable parameters as $\theta = \{W, \bm{b}, \bm{\mu}, \bm{\sigma}^2\}$. To  ensure the variance remains non-negative during learning, we adopt a reparameterization, directly learning $\log \bm{\sigma}^2$ rather than $\bm{\sigma}^2$ or $\bm{\sigma}$.
Finally, given the energy function, one can define the Boltzmann distribution, over visible and hidden states, as
\begin{align}\label{eq:pdf_GRBM}
    p_{\theta}(\mathbf{v}, \mathbf{h}) &= \frac{1}{Z} \exp \left(- E_{\theta}(\mathbf{v}, \mathbf{h}) \right) \, ,~~
    % \\
    % Z &=
    \mbox{where}~~ ~ Z =  \int_{-\infty}^{+\infty} \sum_{\mathbf{h}} \exp \left(- E_{\theta}(\mathbf{v}, \mathbf{h}) \right) \mathrm{d} \mathbf{v}
\end{align}
is the normalization constant, which is intractable for even moderately large $M$.

The underlying graphical model, like an RBM, is a bipartite graph with edges only connecting visible units to hidden units.
This entails conditional independence of the form $p(\mathbf{v} \vert \mathbf{h}) = \prod_i p(\mathbf{v}_i \vert \mathbf{h})$ and $p(\mathbf{h} \vert \mathbf{v}) = \prod_j p(\mathbf{h}_j \vert \mathbf{v})$.
One can also derive the following conditional distributions for GRBMs,
\begin{align}
    p(\mathbf{v} \vert \mathbf{h}) &= \mathcal{N}\left( \mathbf{v} \vert W \mathbf{h} + \bm{\mu}, \mathrm{diag}(\bm{\sigma}^2) \right) \label{eq:v_given_h_GRBM} \\
    p(\mathbf{h}_j = 1 \vert \mathbf{v}) &= \left[ \text{Sigmoid} \left( W^{\top} \frac{\mathbf{v}}{\bm{\sigma}^2} + \mathbf{b} \right) \right]_j, \label{eq:h_given_v_GRBM}
\end{align}
where $\mathcal{N}\left( \mathbf{v} \vert W \mathbf{h} + \bm{\mu}, \mathrm{diag}(\bm{\sigma}^2) \right)$ is the multivariate Gaussian distribution with mean $W \mathbf{h} + \bm{\mu}$, and the diagonal covariance matrix $\mathrm{diag}(\bm{\sigma}^2)$.
Here, $\text{Sigmoid}(\mathbf{x}) = 1/(1+\exp(-\mathbf{x}))$ is applied to the vector $\mathbf{x}$  in an element-wise manner, and
$[\cdot]_j$ denotes the $j$-th element of the corresponding vector.

Given the Boltzmann distribution, one can derive the marginal distribution over visible units, \ie,
\begin{align}\label{eq:marginal_v_GRBM}
     p(\mathbf{v}) &= \frac{1}{Z} \exp\left(-\tilde{E}_{\theta}(\mathbf{v}) \right), \\
     \tilde{E}_{\theta}(\mathbf{v}) &= \frac{1}{2} \left( \frac{\mathbf{v} - \bm{\mu} }{ \bm{\sigma} } \right)^{\top} \left( \frac{\mathbf{v} - \bm{\mu} }{ \bm{\sigma} } \right) - \text{Softplus} \left( W^{\top} \frac{\mathbf{v}}{\bm{\sigma}^2} + \mathbf{b} \right)^{\top} \textbf{1}~, \nonumber
    %  \\
    % p(\mathbf{v}) &= \frac{1}{Z} \exp\left(-\tilde{E}_{\theta}(\mathbf{v}) \right),
\end{align}
where $\text{Softplus}(\mathbf{x}) = \log(1 + \exp(\mathbf{x}))$ is applied  in an element-wise manner, and
$\textbf{1}$ is the all-one vector of  size $M$.
We call $\tilde{E}_{\theta}(\mathbf{v})$  the \textit{marginal energy} to distinguish it from the GRBM energy in Eq. (\ref{eq:energy_GRBM}).
We leave the derivation to Appendix \ref{appendix:marginal_GRBM_derivation}.
As shown in \citet{melchior2017gaussian}, one can also rewrite the marginal distribution $p(\mathbf{v})$ as a constrained Gaussian mixture.

\begin{algorithm}[t]
\caption{Langevin Sampling for GRBMs}
\begin{algorithmic}[1]
  \State \textbf{Input}: $\mathbf{v}^{(0)}$, step size $\alpha_0$, total step $T$, burn-in step $\tilde{T}$, adjust step $\eta$
  \State \textbf{For} $t = 1, \dots, T$
  \State \qquad $\alpha_{t} = \text{CosineScheduler}(t, T, \alpha_0)$
  \State \qquad $\mathbf{v} = \mathbf{v}^{(t-1)} - \alpha_{t} \frac{\partial \tilde{E}(\mathbf{v}^{(t-1)})}{\partial \mathbf{v}} + \sqrt{2 \alpha_{t}} \bm{\xi}_{t}$ ,~~
    %   \quad 
  $\bm{\xi}_{t} \sim \mathcal{N}(0, I)$ \Comment{Use marginal energy in Eq. (\ref{eq:marginal_v_GRBM})}
  \State \qquad \textbf{If} $t <= \eta$ or $\left( t > \eta \text{ and } u \sim \mathcal{U}(0, 1) < A(\mathbf{v}, \mathbf{v}^{(t-1)}) \right)$
  \State \qquad \qquad $\mathbf{v}^{(t)} = \mathbf{v}$
  \State \qquad \textbf{Else}
  \State \qquad \qquad $\mathbf{v}^{(t)} = \mathbf{v}^{(t-1)}$
  \State \textbf{Return}: $\{ \mathbf{v}^{(\tilde{T}+1:T)}\}$ \Comment{$i:j$ indexes consecutive samples from $i$-th to $j$-th}
\end{algorithmic}
\label{alg:Langevin}
\end{algorithm}

% \vspace{-0.1in}
\subsection{Inference}\label{subsec:GRBM_inference}
% \vspace{-0.1in}

When performing probabilistic inference, \eg, computing the marginal distribution or the maximum a posterior (MAP) estimation, one often chooses between variational inference \citep{hinton1993keeping,jordan1999introduction} and Markov chain Monte Carlo (MCMC) methods \citep{neal1993probabilistic,andrieu2003introduction}.
We focus on MCMC as common variational methods have been less effective with RBMs and GRBMs \citep{gabrie2015training,takahashi2016mean}. 
From the generative modelling perspective, we wish to draw samples of visible units during inference.
There are two natural approaches to this: 1) sample from the joint distribution in Eq.\ (\ref{eq:pdf_GRBM}) and discard the samples of hidden units, or 2) directly sample from the marginal distribution.

% In what follows, we describe representative sampling algorithms of these two approaches, and then propose a new hybrid sampler.

% \subsubsection{Gibbs Sampling}\label{subsubsec:Gibbs}

Gibbs sampling \citep{geman1984stochastic} is perhaps the predominant   approach, due to its simplicity.
In the context of GRBMs, one alternates between sampling hidden units given visible units, and sampling visible units given hidden units.
This produces samples from the joint distribution in Eq.\ (\ref{eq:pdf_GRBM}).
The detailed Gibbs sampling algorithm is given in Appendix \ref{appendix:gibbs}.
% Through this Gibbs sampler, we can get samples of visible and hidden units from the joint distribution in Eq. (\ref{eq:pdf_GRBM}).

% \subsubsection{Langevin Sampling}\label{subsubsec:Langevin}
\paragraph{Langevin Sampling}

Langevin Monte Carlo \citep{grenander1994representations,roberts1996exponential,welling2011bayesian} is a class of MCMC methods that generate samples from a probability distribution of continuous random variables by simulating Langevin dynamics.
Since GRBMs are hybrid graphical models, \ie, comprising continuous and discrete random variables, we have at least two ways to leverage Langevin sampling.
One is to directly apply Langevin sampling to the marginal distribution of visible units in Eq.\ (\ref{eq:marginal_v_GRBM}).
Suppose at time step $t-1$, we have sample $\mathbf{v}_{t-1}$ and want to draw a new sample $\mathbf{v}_{t}$.
The proposal distribution corresponding to one-step Langevin dynamics is given by
\begin{align}\label{eq:1step-Langevin}
    q(\mathbf{v} \vert \mathbf{v}^{(t-1)}) = \mathcal{N} \left(\mathbf{v} \middle\vert \mathbf{v}^{(t-1)} - \alpha_{t} \frac{\partial \tilde{E}(\mathbf{v}^{(t-1)})}{\partial \mathbf{v}}, 2 \alpha_{t} I \right),
\end{align}
where the gradient the of marginal energy $\tilde{E}$ \wrt the visible units is given in Appendix \ref{appendix:Langevin_sampling}.
If we use the Metropolis-Hastings algorithm to accept or reject proposed samples, the acceptance probability of a proposal $\mathbf{v}_{t}$, given the previous state,   $\mathbf{v}_{t-1}$, is  (see Appendix \ref{appendix:Langevin_sampling} for derivation):
\small
\begin{align}
    A(\mathbf{v}^{(t)}, \mathbf{v}^{(t-1)}) = \min \left( \! 1,\, \frac{\exp\left( -\tilde{E}_{\theta}(\mathbf{v}^{(t)}) -\frac{1}{4 \alpha_{t}} \left\Vert \mathbf{v}^{(t-1)} - \mathbf{v}^{(t)} + \alpha_{t} \frac{\partial \tilde{E}(\mathbf{v}^{(t)})}{\partial \mathbf{v}} \right\Vert^2 \right) }{\exp\left(-\tilde{E}_{\theta}(\mathbf{v}^{(t-1)}) -\frac{1}{4 \alpha_{t}} \left\Vert \mathbf{v}^{(t)} - \mathbf{v}^{(t-1)} + \alpha_{t} \frac{\partial \tilde{E}(\mathbf{v}^{(t-1)})}{\partial \mathbf{v}} \right\Vert^2 \! \right) } \right) .
\end{align}
\normalsize
% We leave the derivation of this acceptance probability to the appendix.
Alg. \ref{alg:Langevin} shows the Metropolis-adjusted Langevin Algorithm (MALA) for the marginal GRBM.
Compared to generic MALA, it also includes an extra hyperparameter, namely, the adjust step $\eta$.
If $\eta$ is set to $0$, then we perform a Metropolis adjustment at every sampling step, as prescribed in the generic MALA.
If $\eta$ is set to $K > 0$, then we skip the Metropolis adjustment for the first $K$ steps.
The adjust step effectively controls a trade-off between sampling accuracy\footnote{Here sampling accuracy means the closeness between the underlying distribution of samples and the target distribution measured in, \eg, total variation or Wasserstein distances.} and computational efficiency.
Since we do not hope to see Gaussian noise in our final-sampled images (\ie, beyond the level of intrinsic noise in the observations), it is beneficial to decay the noise level, as in score-based models \citep{song2019generative}.
For certain step-size-annealing schedules and energy functions, there are theoretical guarantees on the convergence of Langevin sampling \citep{durmus2019high}.
For simplicity, we use the cosine scheduler and find it works well in practice.
More details about the scheduler are provided in Appendix \ref{appendix:Langevin_sampling}.

% \subsubsection{Gibbs-Langevin Sampling}\label{subsubsec:Gibbs-Langevin}
\paragraph{Gibbs-Langevin Sampling}

\begin{algorithm}[t]
\caption{Gibbs-Langevin Sampling for GRBMs}
\begin{algorithmic}[1]
  \State \textbf{Input}: $\mathbf{v}^{(0)}$, $\mathbf{h}^{(0)}$, step size $\alpha_0$, total step $T$, burn-in step $\tilde{T}$, adjust step $\eta$, Langevin step $K$
  \State \textbf{Function} Langevin($\tilde{\mathbf{v}}^{(0)}$, $\mathbf{h}$, $\alpha_0$, K):
  \State \qquad \textbf{For} $k = 1, \dots, K$
  \State \qquad \qquad $\alpha_{k} = \text{CosineScheduler}(k, K, \alpha_0)$
  \State \qquad \qquad $\tilde{\mathbf{v}}^{(k)} = \tilde{\mathbf{v}}^{(k-1)} - \alpha_{k} \frac{\partial E(\tilde{\mathbf{v}}^{(k-1)}, \mathbf{h})}{\partial \mathbf{v}} + \sqrt{2 \alpha_{k}} \bm{\xi}_{k}$ , ~~
    %   \quad 
  $\bm{\xi}_{k} \sim \mathcal{N}(0, I)$
  \State \textbf{Return} $\tilde{\mathbf{v}}^{(K)}$
  \State
  \State \textbf{For} $t = 1, \dots, T$
  \State \qquad $\mathbf{v} = \text{Langevin}(\mathbf{v}^{(t-1)}, \mathbf{h}^{(t-1)}, \alpha_0, K)$
  \State 
  \qquad $\mathbf{h} \sim p(\mathbf{h} \vert \mathbf{v})$ 
  \State \qquad \textbf{If} $t <= \eta$ or $\left( t > \eta \text{ and } u \sim \mathcal{U}(0, 1) < \tilde{A}\left((\mathbf{v}, \mathbf{h}), (\mathbf{v}^{(t-1)}, \mathbf{h}^{(t-1)})\right) \right)$
  \State \qquad \qquad $\mathbf{v}^{(t)}, \mathbf{h}^{(t)} = \mathbf{v}, \mathbf{h}$
  \State \qquad \textbf{Else}
  \State \qquad \qquad $\mathbf{v}^{(t)}, \mathbf{h}^{(t)} = \mathbf{v}^{(t-1)}, \mathbf{h}^{(t-1)}$
  \State \textbf{Return}: $\{(\mathbf{v}^{(\tilde{T}+1:T)}, \mathbf{h}^{(\tilde{T}+1:T)}) \}$
\end{algorithmic}
\label{alg:Gibbs_Langevin}
\end{algorithm}

We also introduce a new hybrid sampler for GRBMs
% In this section, we propose a new Langevin sampling method for GRBMs 
(see Alg.\ \ref{alg:Gibbs_Langevin}).
Like the Gibbs sampler, it alternates between sampling hidden units conditioned on visible units, and sampling visible units given the hidden units.
% sampling visible units conditioned on hidden ones.
Unlike generic Gibbs, which directly samples from the Gaussian $p(\mathbf{v} \vert \mathbf{h}^{(t)})$, 
we instead use Langevin MC to sample the continuous visible units given the hidden state.
The use of Langevin MC may seem unnecessary because the Gaussian conditional permits a one-step sampling algorithm.
The subtlety comes from the fact that the finite-step Langevin sampler explicitly depends on the initial sample.
% Thus, it allows us to construct a persistent Markov chain in the space of visible units, \ie, starting a sampling trajectory of visible units (conditioned on hidden units) from the latest sampled visible units.
Specifically, the proposal distribution of one complete outer-loop step in Alg.\ \ref{alg:Gibbs_Langevin}, \eg, at iteration $t-1$, can be expressed as
\begin{align}\label{eq:proposal_Gibbs_Langevin}
q(\mathbf{v}, \mathbf{h} \vert \mathbf{v}^{(t-1)}, \mathbf{h}^{(t-1)}) &\, =\, q(\mathbf{h} \vert \mathbf{v})
~q(\mathbf{v} \vert \mathbf{v}^{(t-1)}, \mathbf{h}^{(t-1)}) ~ ,
\end{align}
where $q(\mathbf{h} \vert \mathbf{v})$ is given by Eq.\ (\ref{eq:h_given_v_GRBM}), and $q(\mathbf{v} \vert \mathbf{v}^{(t-1)}, \mathbf{h}^{(t-1)})$ is the proposal distribution of a K-step Langevin sampler (\ie  from the inner loop).
This proposal distribution explicitly depends on the initial visible sample, $\mathbf{v}^{(t-1)}$ from iteration $t\!-\! 1$.
By contrast, the generic  Gibbs sampler does not have such dependence, \ie, $q(\mathbf{v} \vert \mathbf{v}^{(t-1)}, \mathbf{h}^{(t-1)}) = q(\mathbf{v} \vert \mathbf{h}^{(t-1)})$.
This dependence allows us to construct a persistent  Markov chain in the space of visible units. 
Moreover, the Langevin sampler leverages the informative gradient of log density whereas Gibbs sampler does not. 
We find that our new sampler performs significantly better than the vanilla Gibbs sampler in practice.

The Metropolis adjustment for these Gibbs-Langevin proposals is, however, somewhat more involved.
% proposal distribution.
Following Alg.\ \ref{alg:Gibbs_Langevin}, with $\tilde{\mathbf{v}}^{(0)} = \mathbf{v}^{(t-1)}$ and $\tilde{\mathbf{v}}^{(K)} = \mathbf{v}$, by marginalizing out the intermediate states on the Markov chain, we obtain the proposal
\begin{align}\label{eq:proposal_Gibbs_Langevin_inner}
q(\tilde{\mathbf{v}}^{(K)} \vert \tilde{\mathbf{v}}^{(0)}, \mathbf{h}^{(t-1)}) = \int \cdots \int \left( \prod_{k=1}^{K} q(\tilde{\mathbf{v}}^{(k)} \vert \tilde{\mathbf{v}}^{(k-1)}, \mathbf{h}^{(t-1)}) \right) \mathrm{d}\tilde{\mathbf{v}}^{(1)} \cdots \mathrm{d}\tilde{\mathbf{v}}^{(K-1)}. 
\end{align}
The integrand in Eq.\ \ref{eq:proposal_Gibbs_Langevin_inner} comprises $K$ one-step Langevin updates,
% (see Eq.\ (\ref{eq:1step-Langevin})),
each of which is given by
\begin{align}
q(\mathbf{v} \vert \tilde{\mathbf{v}}^{(k-1)}, \mathbf{h}^{(t-1)}) &= \mathcal{N} \left(\mathbf{v} \middle\vert \tilde{\mathbf{v}}^{(k-1)} - \alpha_{k} \frac{\partial E(\tilde{\mathbf{v}}^{(k-1)}, \mathbf{h}^{(t-1)})}{\partial \mathbf{v}},\, 2 \alpha_{k} I \right) ~,
\end{align}
for which the energy gradient is given in Appendix \ref{appendix:Gibbs_Langevin_sampling}.
Although the multiple integral in Eq.\  (\ref{eq:proposal_Gibbs_Langevin_inner}) appears intractable, one can use  reparameterization  to derive the following analytical form,
\small
\begin{align}
   \! q(\tilde{\mathbf{v}}^{(K)} \vert \tilde{\mathbf{v}}^{(0)}, \mathbf{h}^{(t-1)}) = \mathcal{N}\left(\bm{\beta}_0 \tilde{\mathbf{v}}^{(0)} + \left( \sum_{k=1}^{K} \bm{\beta}_{k} \alpha_{k} \right) \frac{\bm{\mu} + W \mathbf{h}^{(t-1)}}{\bm{\sigma}^2}, \, \mathrm{diag}\left( \sum_{k=1}^{K} 2 \alpha_{k} \bm{\beta}_{k}^2 \right) \right),
\end{align}
\normalsize
where $\bm{\beta}_{k} = \prod_{j=k+1}^{K} \left( \bm{1} - \frac{\alpha_{j}}{\bm{\sigma}^2} \right)$, $\forall k \in \{0, \dots, K-1\}$ and $\bm{\beta}_{K} = \bm{1}$.
Based on this result, one can show that the acceptance probability for the Metropolis adjustment is
\small
\begin{align}
    & \tilde{A}((\mathbf{v}^{(t)}, \mathbf{h}^{(t)}), (\mathbf{v}^{(t-1)}, \mathbf{h}^{(t-1)})) = \nonumber \\
    & \min \left(1, 
    \frac{ \exp\left(-E_{\theta}(\mathbf{v}^{(t)}, \mathbf{h}^{(t)}) - \left\Vert \frac{ \mathbf{v}^{(t-1)} - \bm{\beta}_0 \mathbf{v}^{(t)} - \bm{a} (\bm{\mu} + W \mathbf{h}^{(t)} ) }{ \sqrt{2} \tilde{\bm{\sigma}}} \right\Vert^2 \right) q(\mathbf{h}^{(t-1)} \vert \mathbf{v}^{(t-1)}) }
    { \exp\left(-E_{\theta}(\mathbf{v}^{(t-1)}, \mathbf{h}^{(t-1)}) - \left\Vert \frac{ \mathbf{v}^{(t)} - \bm{\beta}_0 \mathbf{v}^{(t-1)} - \bm{a} (\bm{\mu} + W \mathbf{h}^{(t-1)} ) }{ \sqrt{2} \tilde{\bm{\sigma}}} \right\Vert^2 \right) q(\mathbf{h}^{(t)} \vert \mathbf{v}^{(t)}) } 
    \right),
\end{align}
\normalsize
where $q(\mathbf{h}_j = 1 \vert \mathbf{v}) = \left[\text{Sigmoid} \left( W^{\top} \frac{\mathbf{v}}{\bm{\sigma}^2} + \mathbf{b} \right) \right]_{j}$, $\bm{a} = \frac{\sum_{k=1}^{K} \bm{\beta}_{k} \alpha_{k}}{\bm{\sigma}^2}$, and $\tilde{\bm{\sigma}}^2 = \sum_{k=1}^{K} 2 \alpha_{k} \bm{\beta}_{k}^2$.
We leave derivations to Appendix \ref{appendix:Gibbs_Langevin_sampling}.

\subsection{Learning}\label{subsec:learning}

To learn GRBMs, we maximize the log likelihood of the observed data using stochastic gradient-based  methods, \eg, contrastive divergence (CD).
Depending on whether we use the joint (Eq.\ (\ref{eq:pdf_GRBM})) or the marginal (Eq.\ (\ref{eq:marginal_v_GRBM})) distribution, we have two possible  gradient estimators.

\begin{algorithm}[h]
\caption{Modified CD Learning Algorithm for GRBMs with Joint Density}
\begin{algorithmic}[1]
  \State \textbf{Input}: CD-step $K$, burn-in step $M$, learning Rate $\eta$, Langevin step size $\alpha_0$, SGD step $T$
  \State \textbf{For} $t = 1, \cdots, T$
  \State \quad $\mathbf{v}^{+} = \mathbf{v}_{\text{data}}$
  \State \quad $\mathbf{h}^{+} \sim p(\mathbf{h} \vert \mathbf{v}^{+})$ 
  \State \quad $\nabla \theta^{+} = \left\langle \frac{\partial E(\mathbf{v}^{+}, \mathbf{h}^{+})}{\partial \theta} \right\rangle_{d}$ \Comment{Compute Positive Gradient}
  \State \quad $\mathbf{v}^{-}_{0} \sim \mathcal{N}(\mathbf{0}, I)$, $\mathbf{h}^{-}_{0} \sim p(\mathbf{h} \vert \mathbf{v}^{-}_{0})$ \Comment{Modified CD to start with noise}
  \State \quad $\{\mathbf{v}^{-}_{M:K}, \mathbf{h}^{-}_{M:K}\} \sim \text{Sampler}(\mathbf{v}^{-}_{0}, \mathbf{h}^{-}_{0}, \alpha_0 \bar{\bm{\sigma}})$ \Comment{Alg. \ref{alg:Gibbs} or Alg. \ref{alg:Gibbs_Langevin}, $\bar{\bm{\sigma}}$ is current mean variance}
  \State \quad $\nabla \theta^{-} = \left\langle \frac{\partial E(\mathbf{v}^{-}, \mathbf{h}^{-})}{\partial \theta} \right\rangle_{m}$ \Comment{Compute Negative Gradient}
  \State \quad $\theta = \theta - \eta (\nabla \theta^{+} - \nabla \theta^{-})$ \Comment{Compute Update}
  \State \textbf{Return} $\theta$
\end{algorithmic}
\label{alg:learning_joint_GRBM}
\end{algorithm}

\paragraph{Learning with the Joint Distribution}
When optimizing the GRBM with the joint distribution, one can express the general form of the gradient of the log likelihood \wrt parameters $\theta$ as
% First, we introduce the learning algorithm with the joint distribution. One can derive the general-form gradient of the log likelihood \wrt parameters $\theta$ as,
\begin{align}\label{eq:gradients_joint_general}
\nabla \theta = \left\langle -\frac{\partial E_{\theta}(\mathbf{v}, \mathbf{h})}{\partial \theta} \right\rangle_{d} - \left\langle -\frac{\partial E_{\theta}(\mathbf{v}, \mathbf{h})}{\partial \theta}\right\rangle_{m}.
\end{align}
Here, following the notation in the RBM literature, we denote expectation under the \emph{data distribution}, \ie, $p_{\theta}(\mathbf{h} \vert \mathbf{v})\, p_{\text{data}}(\mathbf{v})$, as $\left\langle \cdot \right\rangle_{d} = \mathbb{E}_{p_{\theta}(\mathbf{h} \vert \mathbf{v}) p_{\text{data}}(\mathbf{v})}\left[ \cdot \right]$.
Similarly, we denote the expectation under the \emph{model distribution}, $p_{\theta}(\mathbf{v}, \mathbf{h})$ as $\left\langle \cdot \right\rangle_{m} = \mathbb{E}_{p_{\theta}(\mathbf{v}, \mathbf{h})}\left[ \cdot \right]$.
The expected gradients under the data and model distributions are called \emph{positive} and \emph{negative} gradients respectively.
% in the literature.
Based on Eq.\ (\ref{eq:gradients_joint_general}), we can formulate the gradients of specific parameters as follows,
\begin{align}
    \nabla W_{ij} &= \left\langle \frac{\mathbf{v}_{i}}{\bm{\sigma}_i^2} \mathbf{h}_{j} \right\rangle_{d} - \left\langle \frac{\mathbf{v}_{i}}{\bm{\sigma}_i^2} \mathbf{h}_{j} \right\rangle_{m} \label{eq:gradients_joint_GRBM_1} \\
    \nabla \mu_{i} &= \left\langle \frac{\mathbf{v}_{i} - \mu_{i}}{\bm{\sigma}_i^2} \right\rangle_{d} - \left\langle \frac{\mathbf{v}_{i} - \mu_{i}}{\bm{\sigma}_i^2} \right\rangle_{m} \label{eq:gradients_joint_GRBM_2} \\
    \nabla \log\bm{\sigma}_{i}^2 &= \left\langle \frac{(\mathbf{v}_{i} - \mu_{i})^2}{2\bm{\sigma}_{i}^2} - \frac{\sum_{j} \mathbf{v}_{i} W_{ij} \mathbf{h}_{j}}{\bm{\sigma}_{i}^2} \right\rangle_{d} - \left\langle \frac{(\mathbf{v}_{i} - \mu_{i})^2}{2\bm{\sigma}_{i}^2} - \frac{\sum_{j} \mathbf{v}_{i} W_{ij} \mathbf{h}_{j}}{\bm{\sigma}_{i}^2} \right\rangle_{m} \label{eq:gradients_joint_GRBM_3} \\
    \nabla \mathbf{b}_{i} &= \langle \mathbf{h}_{i} \rangle_{d} - \langle \mathbf{h}_{i} \rangle_{m}  ~. 
    \label{eq:gradients_joint_GRBM_4}
\end{align}
Since the expectations in these gradients are generally intractable, we use Monte Carlo methods to approximate them.
To sample from the joint density, we can use Gibbs or Gibbs-Langevin samplers as described in Sec.\ \ref{subsec:GRBM_inference}.
The overall learning algorithm is outlined in Alg.\ \ref{alg:learning_joint_GRBM}.
An important detail is that we multiply the initial Langevin step size by the average variance at each gradient update step and then feed it to the sampler.
Since the variance is decreasing (the energy function and its gradient are increasing) as learning goes on, keeping the step size roughly invariant to such scaling would make the sampling more effective.

\paragraph{Learning with the Marginal Distribution}
Now we turn to learning the model under the marginal distribution in Eq. (\ref{eq:marginal_v_GRBM}).
Since we have the marginal distribution of visible units, we can directly get the gradients of log likelihood \wrt model parameters $\theta$ as,
\begin{align}\label{eq:gradients_marginal_general}
\nabla \theta = \left\langle -\frac{\partial \tilde{E}_{\theta}(\mathbf{v})}{\partial \theta} \right\rangle_{d} - \left\langle -\frac{\partial \tilde{E}_{\theta}(\mathbf{v})}{\partial \theta}\right\rangle_{m}.
\end{align}
Since the gradient $\frac{\partial \tilde{E}_{\theta}(\mathbf{v})}{\partial \theta}$ does not depend on $\mathbf{h}$ anymore, we have
% \begin{align}
%     \left\langle -\frac{\partial \tilde{E}_{\theta}(\mathbf{v})}{\partial \theta} \right\rangle_{d} &= \mathbb{E}_{p_{\text{data}}(\mathbf{v})} \left[ -\frac{\partial \tilde{E}_{\theta}(\mathbf{v})}{\partial \theta} \right] \nonumber \\
%     \left\langle -\frac{\partial \tilde{E}_{\theta}(\mathbf{v})}{\partial \theta} \right\rangle_{m} &= \mathbb{E}_{p_{\theta}(\mathbf{v})} \left[ -\frac{\partial \tilde{E}_{\theta}(\mathbf{v})}{\partial \theta} \right].
% \end{align}
\begin{align}
    \left\langle -\frac{\partial \tilde{E}_{\theta}(\mathbf{v})}{\partial \theta} \right\rangle_{d} = \mathbb{E}_{p_{\text{data}}(\mathbf{v})} \left[ -\frac{\partial \tilde{E}_{\theta}(\mathbf{v})}{\partial \theta} \right], \qquad
    \left\langle -\frac{\partial \tilde{E}_{\theta}(\mathbf{v})}{\partial \theta} \right\rangle_{m} = \mathbb{E}_{p_{\theta}(\mathbf{v})} \left[ -\frac{\partial \tilde{E}_{\theta}(\mathbf{v})}{\partial \theta} \right].
\end{align}
Based on above results, we can work out the detailed gradients which are the same as those in Eq.\ (\ref{eq:gradients_joint_GRBM_1}) to Eq.\ (\ref{eq:gradients_joint_GRBM_4}) but with $\mathbf{h}$ replaced with $\text{Sigmoid} \left( W^{\top} \frac{\mathbf{v}}{\bm{\sigma}^2} + \mathbf{b} \right)$.
More details are left to Appendix \ref{appendix:learning}.
We use the Langevin sampler in Sec.\ \ref{subsec:GRBM_inference} to sample from the marginal density to approximate the intractable expectation.
The overall learning algorithm is outlined in Alg.\ \ref{alg:learning_marginal_GRBM}.
\begin{algorithm}[t]
\caption{Modified CD Learning Algorithm for GRBMs with Marginal Density}
\begin{algorithmic}[1]
  \State \textbf{Input}: CD-step $K$, burn-in step $M$, learning rate $\eta$, Langevin step size $\alpha_0$, SGD step $T$
  \State \textbf{For} $t = 1, \cdots, T$
  \State \quad $\mathbf{v}^{+} = \mathbf{v}_{\text{data}}$
  \State \quad $\nabla \theta^{+} = \left\langle \frac{\partial \tilde{E}(\mathbf{v}^{+})}{\partial \theta} \right\rangle_{d}$ \Comment{Compute Positive Gradient}
  \State \quad $\mathbf{v}^{-}_{0} \sim \mathcal{N}(\mathbf{0}, I)$ \Comment{Modified CD to start with noise}
  \State \quad $\{\mathbf{v}^{-}_{i} \vert i = M, \cdots, K\} \sim \text{Sampler}(\mathbf{v}^{-}_{0}, \alpha_0 \bar{\bm{\sigma}})$ \Comment{Alg. \ref{alg:Langevin}, $\bar{\bm{\sigma}}$ is current mean variance}
  \State \quad $\nabla \theta^{-} = \left\langle \frac{\partial \tilde{E}(\mathbf{v}^{-})}{\partial \theta} \right\rangle_{m}$ \Comment{Compute Negative Gradient}
  \State \quad $\theta = \theta - \eta (\nabla \theta^{+} - \nabla \theta^{-})$ \Comment{Compute Update}
  \State \textbf{Return} $\theta$
\end{algorithmic}
\label{alg:learning_marginal_GRBM}
\end{algorithm}

\paragraph{Modified Contrastive Divergence}
The above two learning algorithms resemble CD if one ignores the specific sampler used.
There exists a subtle yet important difference however.
% In CD, one runs a single Markov chain, starting from observed data, to collect samples for both positive and negative gradients.
% The underlying intuition is that if the learned model is good, the Markov chain should leave the data distribution invariant.
% Therefore, the Markov chain should behave similarly whether starting from observed data (positive phase) or reconstructed data\footnote{The reconstructed data is typically obtained by running one complete step of Gibbs sampler from the observed data, thus being highly likely close to observed data.} (negative phase).
% This intuition is useful for in learning.
% However, from the perspective of density modelling, it entails challenges in testing a learned model, since we cannot start the Markov chain (sampling) from some observed data during testing.
% From the perspective of generative modeling, this creates challenges for testing the model.
For most deep generative models one generates samples
starting from noise.
% like standard Normal during testing. 
But this does not work well for models trained with CD,
where sampling starts from observed data.
% in CD during training.
This discrepancy of the starting sample between training and testing would be a significant issue if the Markov chain does not mix sufficiently quickly.
We therefore modify CD by running two Markov chains to collect samples for positive and negative gradients respectively.
The positive Markov chain is the same as in CD, \ie, starting from observed data.
The negative Markov chain now starts from a sample of standard Normal noise rather than the reconstructed data\footnote{The reconstructed data is typically obtained by running one complete step of Gibbs sampler from the observed data, thus being highly likely close to observed data.}.
Since the positive chain starting from data will usually stay close to the data distribution, this modification pushes the negative Markov chain, starting from noise, toward the data distribution.
% This intuition will likely to be the case if the model is learning well and the negative chain runs for enough steps.  
Moreover, the discrepancy between training and testing ceases to be important as we can start from standard Normal noise while sampling from the learned model.

\section{Experiments}\label{sec:exp}

We examine the empirical behavior of our new GRBM algorithms on benchmark image datasets, namely, MNIST, Fashion-MNIST \citep{xiao2017online}, and CelebA \citep{liu2015faceattributes}.

\begin{figure}[t]
\centering
\begin{subfigure}{.99\textwidth}
    \centering
    \begin{tabular}{@{}l@{}l@{}l@{}l@{}l@{}l@{}}
    \includegraphics[width=.162\linewidth]{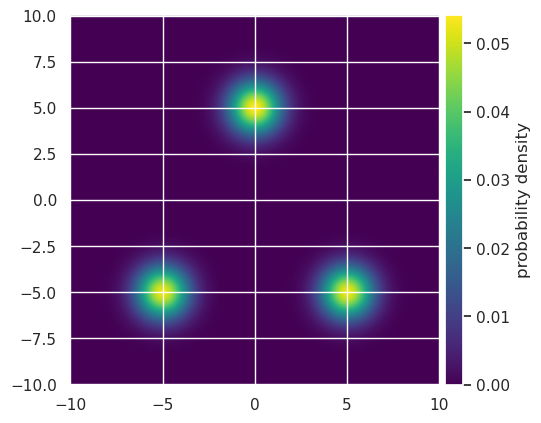} 
    &
    \includegraphics[width=.16\linewidth]{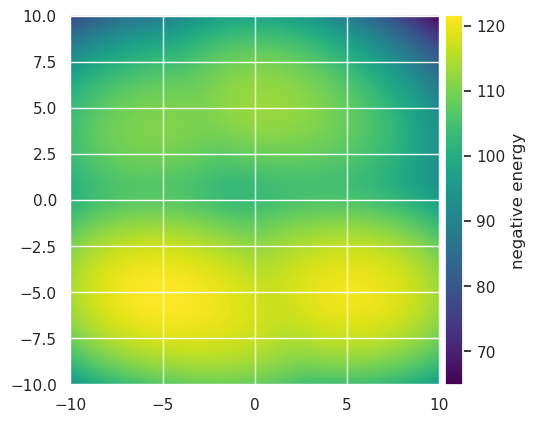} 
    &
    \includegraphics[width=.17\linewidth]{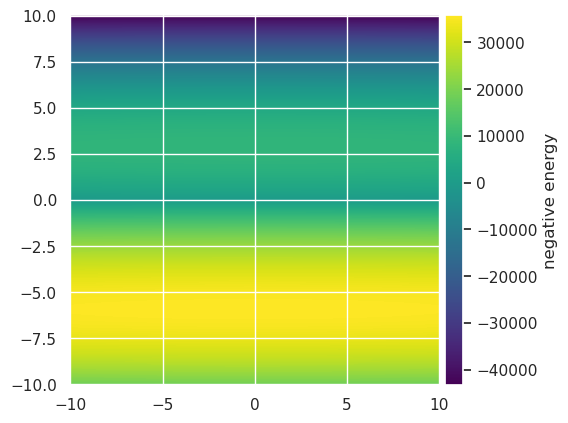} 
    &
    \includegraphics[width=.16\linewidth]{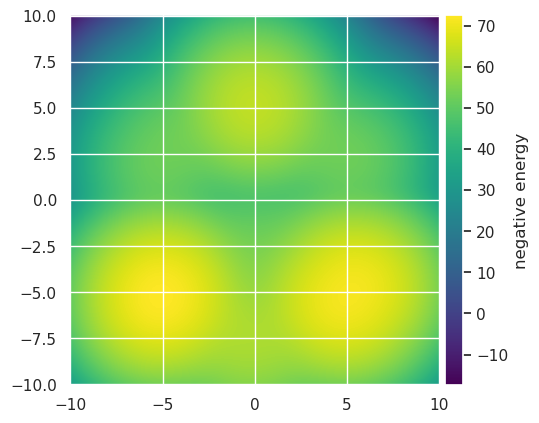} 
    &
    \includegraphics[width=.17\linewidth]{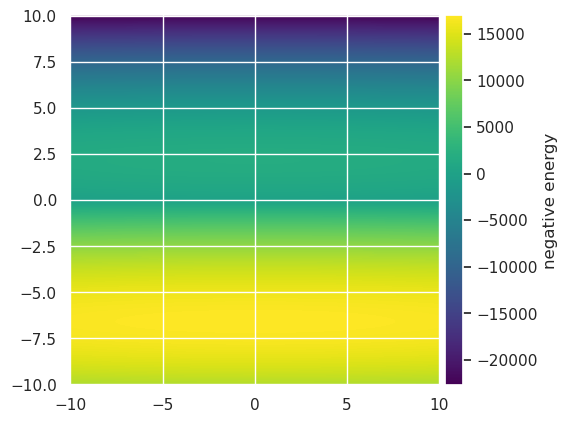} 
    &
    \includegraphics[width=.16\linewidth]{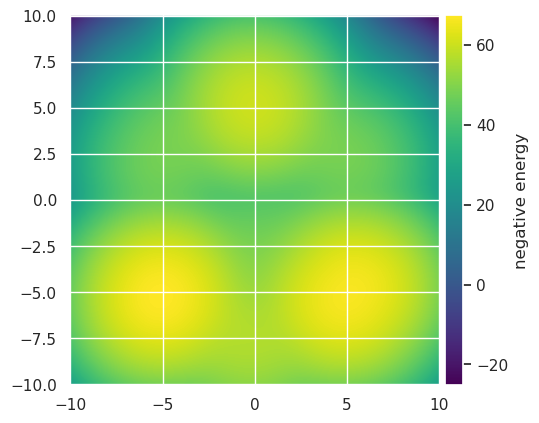}
    \\
    \includegraphics[width=.14\linewidth]{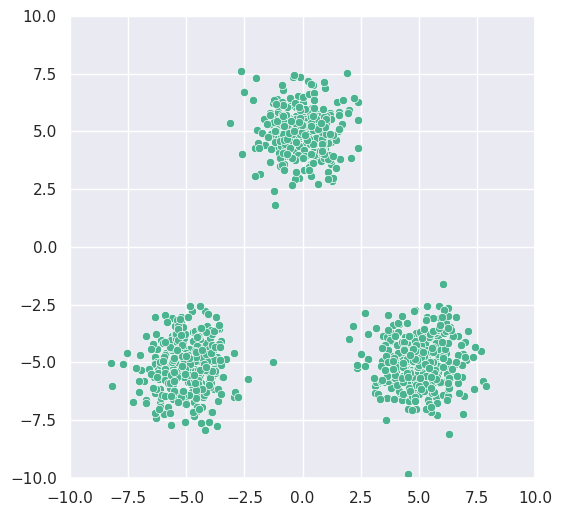}
    &
    \includegraphics[width=.14\linewidth]{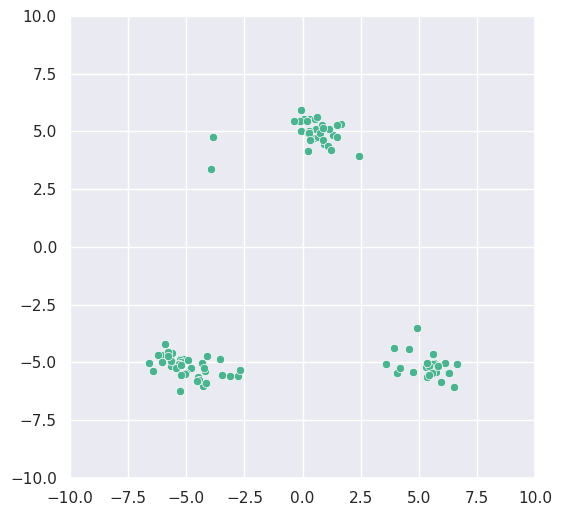}  
    &
    \includegraphics[width=.14\linewidth]{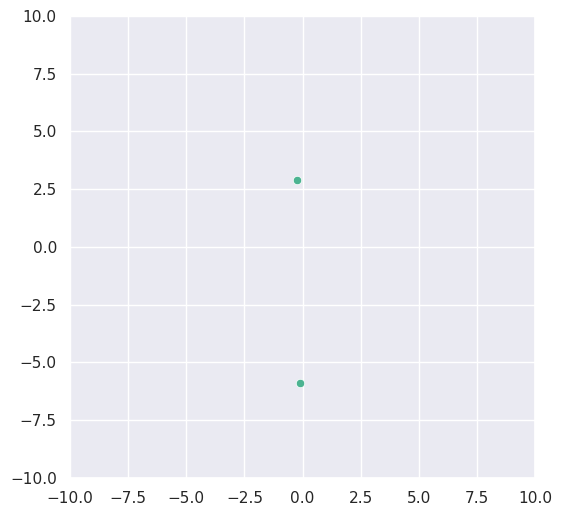}  
    &
    \includegraphics[width=.14\linewidth]{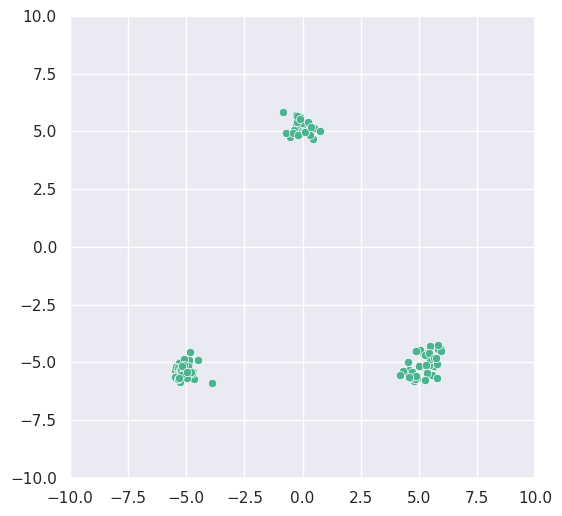}  
    &
    \includegraphics[width=.14\linewidth]{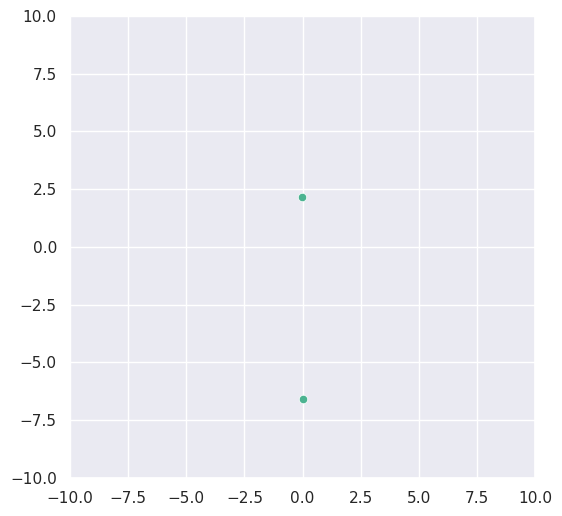}  
    &
    \includegraphics[width=.14\linewidth]{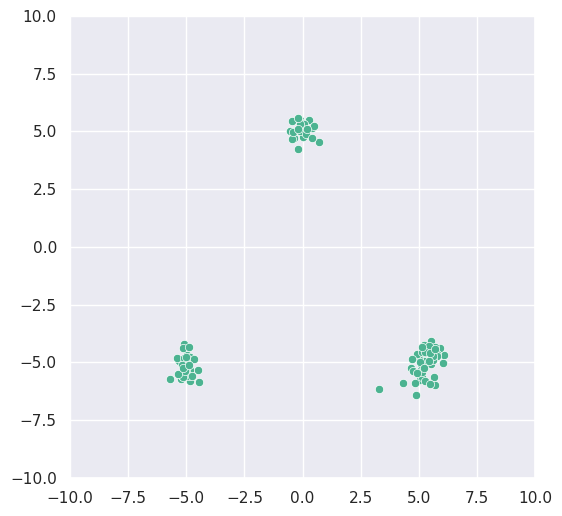}    
    \\
    \includegraphics[width=.162\linewidth]{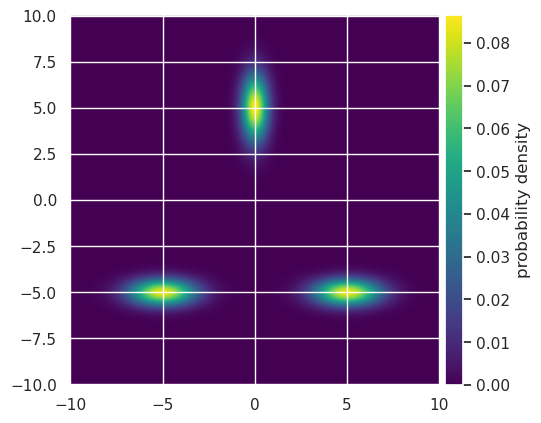}  
    &
    \includegraphics[width=.16\linewidth]{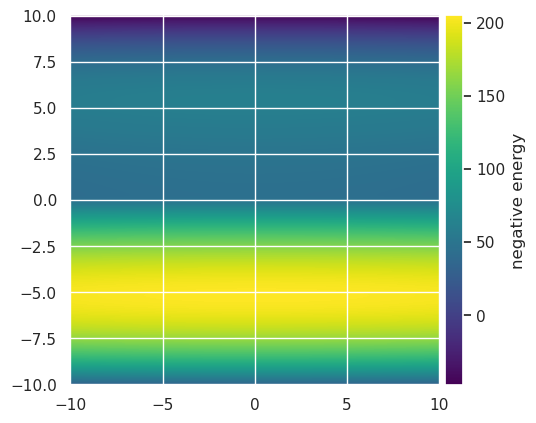}  
    &
    \includegraphics[width=.17\linewidth]{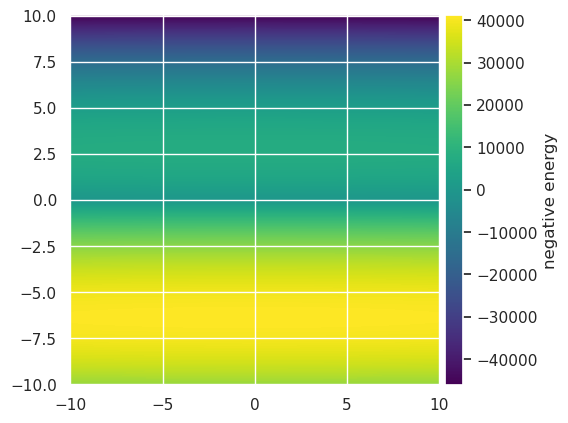}  
    &
    \includegraphics[width=.16\linewidth]{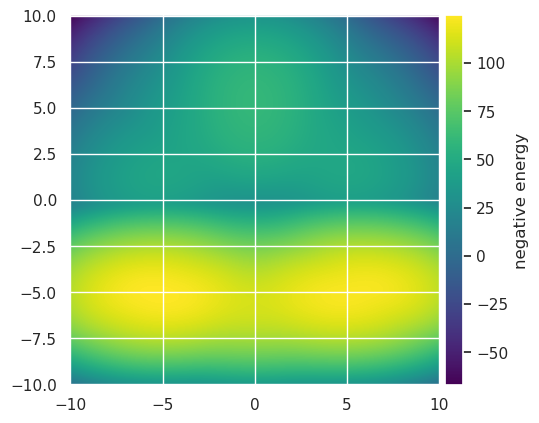}  
    &
    \includegraphics[width=.17\linewidth]{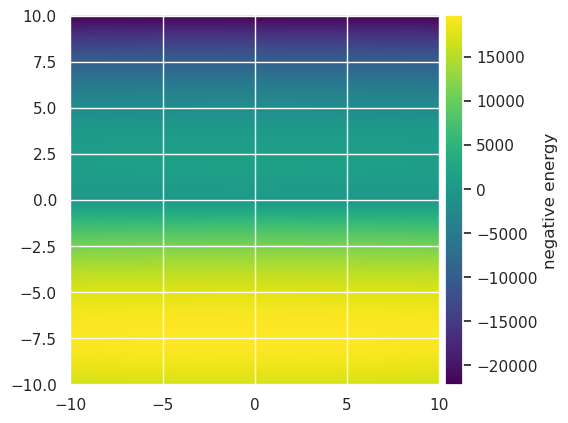}  
    &
    \includegraphics[width=.16\linewidth]{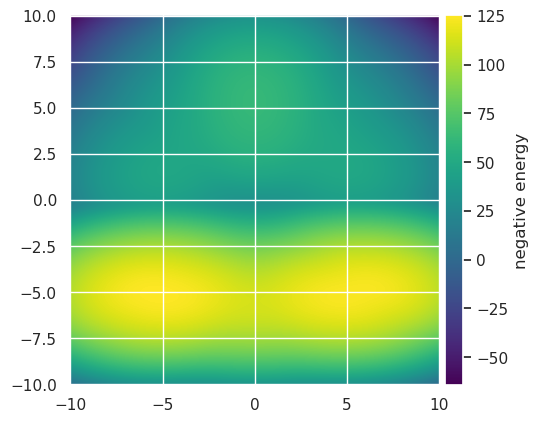}
    \\    
    \includegraphics[width=.14\linewidth]{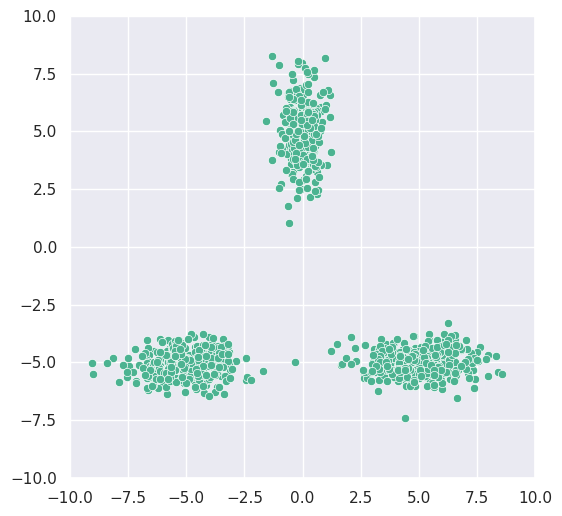} 
    &
    \includegraphics[width=.14\linewidth]{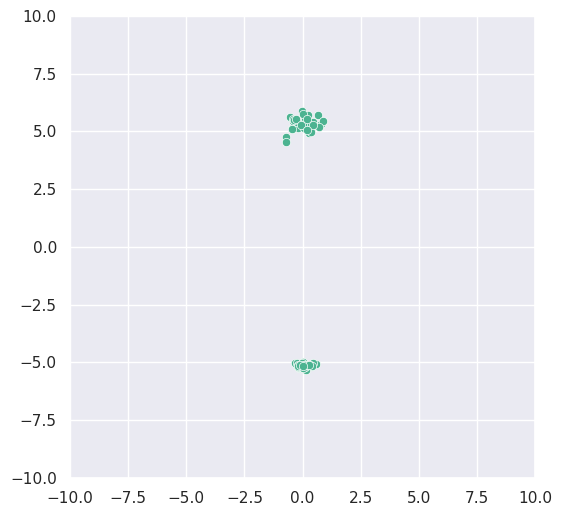} 
    &
    \includegraphics[width=.14\linewidth]{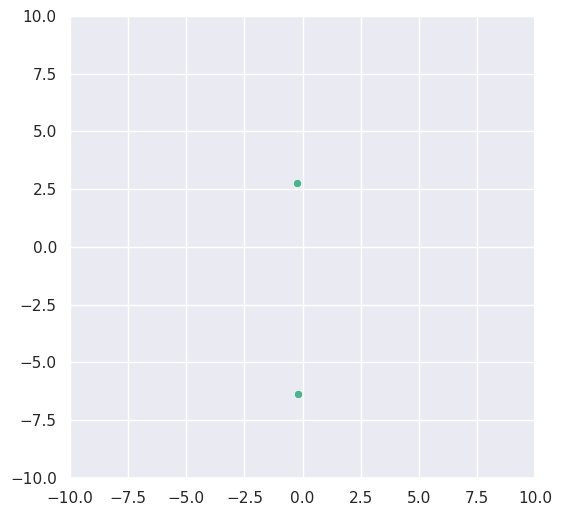} 
    &
    \includegraphics[width=.14\linewidth]{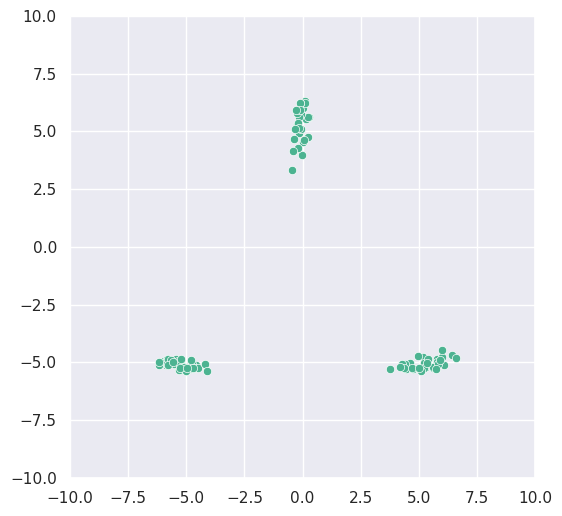} 
    &
    \includegraphics[width=.14\linewidth]{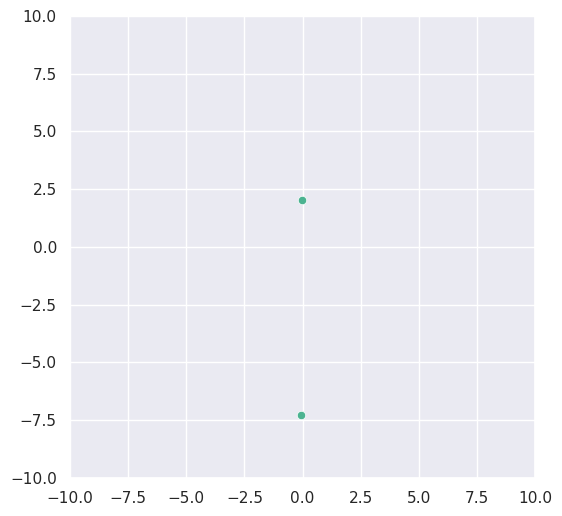} 
    &
    \includegraphics[width=.14\linewidth]{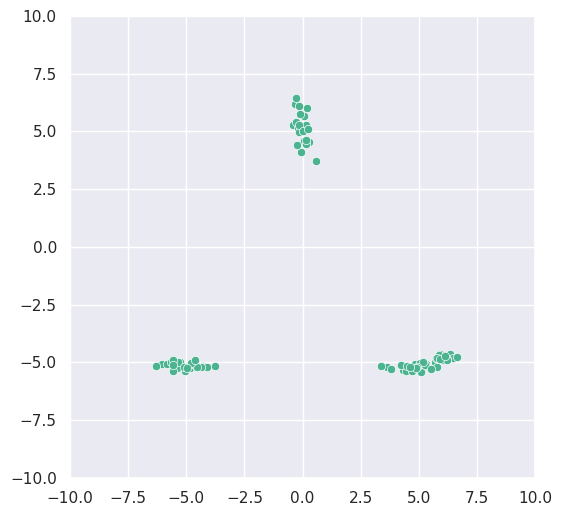}
    \\
    \multicolumn{1}{c}{(a)} & \multicolumn{1}{c}{(b)} & \multicolumn{1}{c}{(c)} & \multicolumn{1}{c}{(d)} & \multicolumn{1}{c}{(e)} & \multicolumn{1}{c}{(f)}
    \end{tabular}
\end{subfigure}
\vspace{-0.1in}
\caption{
Density modelling using GRBMs on data from a Gaussian mixtures with isotropic (rows 1 and 2) and anisotropic variances (rows 3 and 4). 
Rows 1 and 3 show normalized GMM densities and (unnormalized) negative energy values for GRBMs. 
Rows 2 and 4 show samples drawn under different models and methods; \ie,
(a) Ground Truth; (b) Gibbs; (c) Langevin wo.\ Adjust; (d) Langevin w.\ Adjust; (e) Gibbs-Langevin wo.\ Adjust; (f) Gibbs-Langevin w.\ Adjust.
}
\vspace{-0.2in}
\label{fig:gmm}
\end{figure}

\paragraph{Implementation Details} We found that training with modified CD alone occasionally diverges, necessitating careful tuning of the learning rate. 
However, adding gradient clipping (\eg, clip gradient norm to 10) enables stable training with all aforementioned sampling methods.
We therefore set learning rate to $0.01$ for all experiments.
Such a large learning rate almost never works in the literature. 
\citet{melchior2017gaussian} used gradient clipping and similarly large learning rates, but they had to set the learning rate for the variances 100 times smaller than that for the weights and biases during CD training.
But thanks to the modified CD and gradient clipping, we found this special treatment of variances is unnecessary.
We do not use momentum, weight decay, PCD, or other tricks.

\subsection{Modeling Gaussian Mixture Densities}

We first evaluate density modelling by GRBMs when the data density is known, \ie, Gaussian mixture models (GMMs) in our case.
This is challenging  for GRBMs as the marginal distribution of visible units of GRBMs is essentially a constrained Gaussian mixture, \ie, the weights of mixture components depend on one another \citep{melchior2017gaussian}.
As such, the mixture components in GRBMs can not be freely placed in the visible domain so one actually needs more hidden units than the log of the number of mixture components to fit GMMs well.
We consider the 2D case for simplicity and better visibility.
We generate 1,000 samples from two types (isotropic and anisotropic variances) of GMMs with $3$ components as shown in Fig. \ref{fig:gmm}, and learn GRBMs using our modified CD with different sampling algorithms,
from which we can draw samples. 
Here all samplers run for 100 steps during both CD training and testing (see Appendix \ref{appendix:GMM} for more detail).
Density plots and samples are shown in Fig.\ \ref{fig:gmm}.
Notice that Gibbs manages to recover the three modes in the isotropic case but fails in the anisotropic case.
Both Langevin and Gibbs-Langevin sampling collapse when the adjustment is absent. 
We believe the cosine step size schedule contributes to the collapse as it removes more stochasticity of Langevin dynamics with small step sizes, thus making sampling more similar to gradient descent.
But as we will see later, in image modelling, this may not be so severe; there are more modes so that the sampling may collapse to different modes, and the diversity of images remains acceptable. 
Finally, both Langevin and Gibbs-Langevin do recover all three modes with the adjustment, which shows the adjustment helps the mixing in this synthetic case.

\subsection{Image Generation}

We learn GRBMs to fit image datasets including MNIST, FashionMNIST, and CelebA.
To the best of our knowledge, this is the first time that GRBMs have been shown to (unconditionally) generate good images. We provide the ablation study in Appendix~\ref{appendix:ablation} and more results in Appendix \ref{appendix:more_vis}.
%More details about experiments and the ablation study are left to the appendix.

\begin{minipage}{\textwidth}
    \begin{minipage}[b]{0.49\textwidth}
        \centering
        \renewcommand*{\arraystretch}{0}
        \begin{tabular}{@{}c@{}c@{}c@{}c@{}c@{}c@{}c@{}c@{}}
        \includegraphics[width=.124\linewidth]{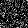} 
        &
        \includegraphics[width=.124\linewidth]{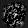} 
        &
        \includegraphics[width=.124\linewidth]{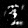} 
        &
        \includegraphics[width=.124\linewidth]{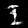} 
        &
        \includegraphics[width=.124\linewidth]{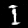} 
        &    
        \includegraphics[width=.124\linewidth]{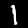} 
        &        
        \includegraphics[width=.124\linewidth]{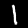} 
        &
        \includegraphics[width=.124\linewidth]{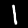}
        \\
        \includegraphics[width=.124\linewidth]{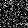} 
        &
        \includegraphics[width=.124\linewidth]{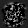} 
        &
        \includegraphics[width=.124\linewidth]{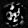} 
        &
        \includegraphics[width=.124\linewidth]{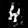} 
        &
        \includegraphics[width=.124\linewidth]{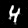} 
        &    
        \includegraphics[width=.124\linewidth]{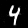} 
        &        
        \includegraphics[width=.124\linewidth]{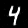} 
        &
        \includegraphics[width=.124\linewidth]{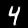}
        \\
        \includegraphics[width=.124\linewidth]{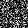} 
        &
        \includegraphics[width=.124\linewidth]{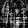} 
        &
        \includegraphics[width=.124\linewidth]{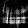} 
        &
        \includegraphics[width=.124\linewidth]{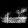} 
        &
        \includegraphics[width=.124\linewidth]{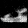} 
        &    
        \includegraphics[width=.124\linewidth]{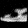} 
        &        
        \includegraphics[width=.124\linewidth]{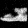} 
        &
        \includegraphics[width=.124\linewidth]{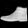}
        \\
        \includegraphics[width=.124\linewidth]{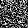} 
        &
        \includegraphics[width=.124\linewidth]{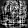} 
        &
        \includegraphics[width=.124\linewidth]{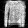} 
        &
        \includegraphics[width=.124\linewidth]{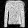} 
        &
        \includegraphics[width=.124\linewidth]{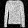} 
        &    
        \includegraphics[width=.124\linewidth]{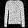} 
        &        
        \includegraphics[width=.124\linewidth]{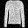} 
        &
        \includegraphics[width=.124\linewidth]{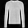}
        \\
        \includegraphics[width=.124\linewidth]{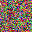} 
        &
        \includegraphics[width=.124\linewidth]{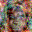} 
        &
        \includegraphics[width=.124\linewidth]{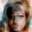} 
        &
        \includegraphics[width=.124\linewidth]{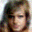} 
        &
        \includegraphics[width=.124\linewidth]{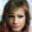} 
        &    
        \includegraphics[width=.124\linewidth]{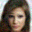} 
        &        
        \includegraphics[width=.124\linewidth]{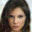} 
        &
        \includegraphics[width=.124\linewidth]{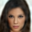}
        \\
        \includegraphics[width=.124\linewidth]{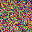} 
        &
        \includegraphics[width=.124\linewidth]{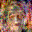} 
        &
        \includegraphics[width=.124\linewidth]{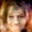} 
        &
        \includegraphics[width=.124\linewidth]{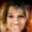} 
        &
        \includegraphics[width=.124\linewidth]{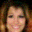} 
        &    
        \includegraphics[width=.124\linewidth]{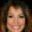} 
        &        
        \includegraphics[width=.124\linewidth]{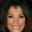} 
        &
        \includegraphics[width=.124\linewidth]{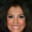}
        \end{tabular}
        \vspace{-0.1in}
        \captionof{figure}{Intermediate samples from Gibbs-Langevin sampling.}
        \label{fig:sampling}
    \end{minipage}
    \hfill
    % \begin{minipage}[b]{0.49\textwidth}
    %     \centering
    %     {%
    %     \begin{tabular}{@{}l|cc@{}}
    %     \hline
    %     \toprule
    %     Methods & FID & IS \\
    %     \midrule
    %     \midrule
    %     VAE & 16.13 & - \\
    %     2sVAE \citep{dai2019diagnosing} & 12.60 & - \\
    %     PixelCNN++ \citep{salimanspixelcnn++} & 11.38 & - \\
    %     WGAN \citep{arjovsky2017wasserstein} & 10.28 & - \\
    %     % WGAN-GP & & \\
    %     NVAE \citep{vahdat2020nvae} & 7.93 & - \\
    %     \midrule
    %     GRBMs &  & \\
    %     \midrule
    %     Gibbs & 47.53  & 2.363 \\
    %     Langevin wo. Adjust & 43.80 & 2.430 \\
    %     Langevin w. Adjust & 41.24 & 2.379 \\
    %     Gibbs-Langevin wo. Adjust & 17.49 & 2.432 \\
    %     Gibbs-Langevin w. Adjust & 19.27 & 2.414 \\
    %     \bottomrule
    %     \end{tabular}
    %     }
    %     \vspace{-0.1in}
    %     \captionof{table}{Results on MNIST dataset.}
    %     \label{tbl:MNIST}
    % \end{minipage}
    \begin{minipage}[b]{0.49\textwidth}
        \centering
        {%
        \begin{tabular}{@{}l|c@{}}
        \hline
        \toprule
        Methods & FID \\
        \midrule
        \midrule
        VAE & 16.13 \\
        2sVAE \citep{dai2019diagnosing} & 12.60 \\
        PixelCNN++ \citep{salimanspixelcnn++} & 11.38 \\
        WGAN \citep{arjovsky2017wasserstein} & 10.28 \\
        % WGAN-GP & & \\
        NVAE \citep{vahdat2020nvae} & 7.93 \\
        \midrule
        GRBMs &  \\
        \midrule
        Gibbs & 47.53 \\
        Langevin wo.\ Adjust & 43.80 \\
        Langevin w.\ Adjust & 41.24 \\
        Gibbs-Langevin wo.\ Adjust & 17.49 \\
        Gibbs-Langevin w.\ Adjust & 19.27 \\
        \bottomrule
        \end{tabular}
        }
        \vspace{-0.1in}
        \captionof{table}{Results on MNIST dataset.}
        \label{tbl:MNIST}
    \end{minipage}
\end{minipage}
\vspace{-0.1in}

\begin{figure}[t]
\centering
\begin{subfigure}{.4\textwidth}
    \centering
    \includegraphics[width=.99\linewidth]{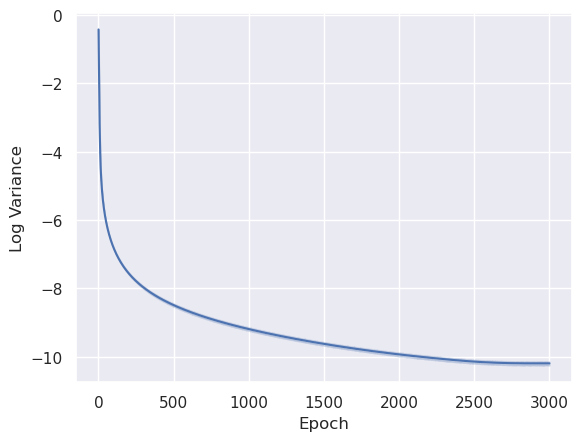}  
    \vspace{-0.2in}
    \caption{}
    \label{fig:varianceMNIST}
\end{subfigure}
\begin{subfigure}{.29\textwidth}
    \centering
    \includegraphics[width=.99\linewidth]{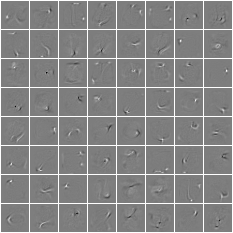}  
    \vspace{-0.2in}
    \caption{}
    \label{fig:filtersMNIST}
\end{subfigure}
\begin{subfigure}{.29\textwidth}
    \centering
    \includegraphics[width=.99\linewidth]{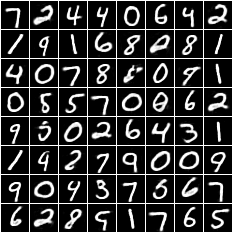}
    \vspace{-0.2in}
    \caption{}
    \label{fig:MNIST}    
\end{subfigure}
\vspace{-0.1in}
\caption{(a) Learning curve of (natural) log variances, (b) learned filters, and (c) samples on MNIST.}
\vspace{-0.1in}
\end{figure}

\paragraph{MNIST} 
We train GRBMs with hidden size 4096 and 100 sampling steps on MNIST.
We compare FID scores of GRBMs with other deep generative models in Table \ref{tbl:MNIST}.
From the table, we can see that Gibbs-Langevin family works significantly better than the Langevin family.
The Metropolis adjustment improves Langevin slightly but degrades Gibbs-Langevin slightly, which is different from what we observed on synthetic data.
This is likely because the image distribution is so complicated (\eg, having significantly more modes) that the adjustment rejects proposed moves more frequently than before.
Some sophisticated strategy may be needed to increase the acceptance probability.
Nevertheless, GRBMs trained with Gibbs-Langevin without adjustment achieve FID scores comparable to other deep generative models, which is impressive given the single-hidden-layer architecture.
The learning curve of (natural) log variance is shown in Fig. \ref{fig:varianceMNIST}.
The learned variance converges to around $1e^{-5}$ which is significantly smaller than those reported in the literature.
The learned filters are shown in Fig. \ref{fig:filtersMNIST}.
Although some point-like filters still exist, stroke-like filters are common, thus indicating GRBMs indeed learn meaningful features.
We show samples drawn from the best GRBM in Fig. \ref{fig:MNIST}.
The intermediate samples from Gibbs-Langevin are shown in Fig. \ref{fig:sampling}.
Since Gibbs-Langevin without adjustment works the best, we use it for remaining experiments.

\begin{figure}[t]
\centering
\begin{subfigure}{.325\textwidth}
    \centering
    \includegraphics[width=.99\linewidth]{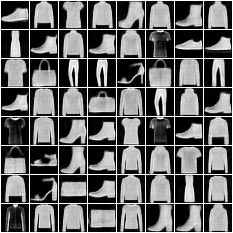}
    \vspace{-0.2in}
    \caption{}
    \label{fig:FashionMNIST}
\end{subfigure}
\begin{subfigure}{.325\textwidth}
    \centering
    \includegraphics[width=.99\linewidth]{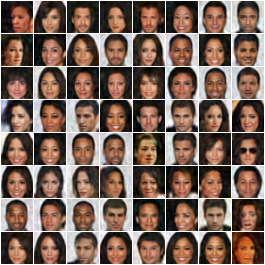}  
    \vspace{-0.2in}
    \caption{}
    \label{fig:CelebA32}    
\end{subfigure}
\begin{subfigure}{.325\textwidth}
    \centering
    \includegraphics[width=.99\linewidth]{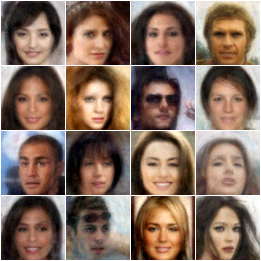}  
    \vspace{-0.2in}
    \caption{}
    \label{fig:CelebA2K64}    
\end{subfigure}
\vspace{-0.1in}
\caption{
Samples from GRBMs on (a) FashionMNIST, (b) CelebA-32, and (c) CelebA-2K-64.
% Samples from GRBMs on (a) FashionMNIST, (b) CelebA-32 (32 $\!\times\!$ 32), and (c) CelebA-2K-64 (64 $\!\times\!$ 64).
}
\vspace{-0.2in}
\end{figure}

\paragraph{FashionMNIST} 
We then train GRBMs on FahsionMNIST which is more challenging than MNIST.
We set hidden size to 10,000 and the sampling step to 100.
Samples drawn from learned GRBMs are shown in Fig. \ref{fig:FashionMNIST}.
GRBMs successfully learn the shapes of clothes, shoes, bags, and so on.
However, they fail to capture fine textures. Since many images in this dataset look similar in shape but differ in texture, the resulting samples look similar to each other. 

\paragraph{CelebA}
Last, we consider the even more challenging CelebA dataset.
In particular, we explore two versions of this dataset: 1) CelebA-32 where we center-crop (140 $\!\times\!$ 140) and downsample images to 32 $\!\times\!$ 32; 2) CelebA-2K-64 where randomly select 2,000 images from the original CelebA and apply the same center crop and downsampling to 64 $\!\times\!$ 64.
We set hidden size to 10,000 and explore the number of 100 and 200 sampling steps.
Generated samples are shown in Fig. \ref{fig:CelebA32} and \ref{fig:CelebA2K64}.
From the figure, we can see that GRBMs can learn to generate reasonably good face images.

\section{Conclusion}\label{sec:conclusion}

In this paper, we revisit learning Gaussian-Bernoulli restricted Boltzmann machines.
We investigate Langevin Monte Carlo and propose a novel Gibbs-Langevin sampling method.
Furthermore, we modify the contrastive divergence (CD) algorithm so that one can sample data from learned GRBMs starting from noise.
Modified CD along with gradient clipping enables robust training of GRBMs with large learning rates.
Finally, we show that GRBMs can unconditionally generate images with good qualities, despite its single-hidden-layer architecture.
In the future, it would be beneficial to extend the current GRBMs to convolutional GRBMs which should be able to learn better localized filters.
Meanwhile, it would be interesting to explore Gaussian deep belief networks (GDBNs), which are deeper than GRBMs and should be superior.
At last, investigating our Gibbs-Langevin sampling for hybrid deep energy based models could be a fruitful direction.

% \subsubsection*{Author Contributions}
% If you'd like to, you may include  a section for author contributions as is done
% in many journals. This is optional and at the discretion of the authors.

\subsubsection*{Acknowledgments}
This work was funded, in part, by the NSERC Discovery Grant.
Resources used in preparing this research were provided, in part, by Google, Vector Institute, and UBC. 

% \clearpage

\bibliography{references}
\bibliographystyle{iclr2023_conference}

\clearpage
\appendix
\section{Derivations}

\subsection{Marginal Probability Distribution of Visible Units of GRBMs}\label{appendix:marginal_GRBM_derivation}
We derive the marginal distribution of visible units as follows,
\begin{align}\label{eq:marginal_v_GRBM_full}
    p(\mathbf{v}) &= \sum_{\mathbf{h}} p(\mathbf{v}, \mathbf{h})  \nonumber \\
    &= \frac{1}{Z} \sum_{\mathbf{h}} \exp(-E_{\theta}(\mathbf{v}, \mathbf{h})) \nonumber \\
    &= \frac{1}{Z} \exp\left(-\frac{1}{2} \left( \frac{\mathbf{v} - \bm{\mu} }{ \bm{\sigma} } \right)^{\top} \left( \frac{\mathbf{v} - \bm{\mu} }{ \bm{\sigma} } \right)\right) \sum_{\mathbf{h}} \exp\left(\left( \frac{\mathbf{v}}{\bm{\sigma}^2} \right)^{\top} W \mathbf{h} + \mathbf{b}^{\top} \mathbf{h}\right) \nonumber \\
    &= \frac{1}{Z} \exp\left(-\frac{1}{2} \left( \frac{\mathbf{v} - \bm{\mu} }{ \bm{\sigma} } \right)^{\top} \left( \frac{\mathbf{v} - \bm{\mu} }{ \bm{\sigma} } \right)\right) \prod_{i} \left( 1 + \exp\left( \left( \left( \frac{\mathbf{v}}{\bm{\sigma}^2} \right)^{\top}W \right)_{i} + \mathbf{b}_{i} \right) \right) \nonumber \\
    &= \frac{1}{Z} \exp\left(-\frac{1}{2} \left( \frac{\mathbf{v} - \bm{\mu} }{ \bm{\sigma} } \right)^{\top} \left( \frac{\mathbf{v} - \bm{\mu} }{ \bm{\sigma} } \right)\right) \prod_{i} \exp \left( \text{Softplus} \left( \left( \left( \frac{\mathbf{v}}{\bm{\sigma}^2} \right)^{\top}W \right)_{i} + \mathbf{b}_{i} \right)\right) \nonumber \\
    &= \frac{1}{Z} \exp\left(-\frac{1}{2} \left( \frac{\mathbf{v} - \bm{\mu} }{ \bm{\sigma} } \right)^{\top} \left( \frac{\mathbf{v} - \bm{\mu} }{ \bm{\sigma} } \right)\right) \exp \left( \text{Softplus} \left( W^{\top} \frac{\mathbf{v}}{\bm{\sigma}^2} + \mathbf{b} \right)^{\top} \textbf{1} \right) \nonumber \\
    &= \frac{1}{Z} \exp\left(-\frac{1}{2} \left( \frac{\mathbf{v} - \bm{\mu} }{ \bm{\sigma} } \right)^{\top} \left( \frac{\mathbf{v} - \bm{\mu} }{ \bm{\sigma} } \right) + \text{Softplus} \left( W^{\top} \frac{\mathbf{v}}{\bm{\sigma}^2} + \mathbf{b} \right)^{\top} \textbf{1} \right).
\end{align}

\subsection{Gibbs Sampling}\label{appendix:gibbs}

\begin{algorithm}
\caption{Gibbs Sampling for GRBMs}
\begin{algorithmic}[1]
  \State \textbf{Input}: number of steps $T$, burn-in step $\tilde{T}$
  \State $\mathbf{v}^{(0)} \sim \mathcal{N}(\mathbf{0}, I)$
  \State \textbf{For} $t = 1, \dots, T$
  \State \qquad $\mathbf{h}^{(t)} \sim p(\mathbf{h} \vert \mathbf{v}^{(t-1)})$ \Comment{following Eq.\ (\ref{eq:h_given_v_GRBM})}
  \State \qquad $\mathbf{v}^{(t)} \sim p(\mathbf{v} \vert \mathbf{h}^{(t)})$ \Comment{following Eq.\ (\ref{eq:v_given_h_GRBM})}
  \State \textbf{Return}: $\{(\mathbf{v}^{(t)}, \mathbf{h}^{(t)}) \vert t = \tilde{T}+1, \cdots, T\}$
\end{algorithmic}
\label{alg:Gibbs}
\end{algorithm}

Gibbs sampling \citep{geman1984stochastic} is perhaps the most popular approach due to its simplicity.
In the context of GRBMs, we can alternate between sampling hidden units given visible units and sampling visible units given hidden units.
% The specific algorithm is shown in 
Alg.\ \ref{alg:Gibbs} is a 
% Note that this is actually a 
blocked Gibbs sampler; it samples all visible units (a block of random variables) at once given all hidden units (the other block) and vice versa.
Given the  conditional independence in the bipartite graphical model, this block Gibbs sampler is equivalent to a  univariate Gibbs sampler that updates one variable at a time given the others following some schedule.
In fact, any schedule comprising a sequence of all hidden units followed by all visible units or vice versa would make the equivalence hold.
In other words, it preserves the convergence of the original univariate Gibbs sampler and runs as fast as a blocked Gibbs sampler.
Relying on this Gibbs sampler, we can get samples of visible and hidden units from the joint distribution in Eq. (\ref{eq:pdf_GRBM}).
We can then discard the samples within the burn-in stage and treat remaining ones as the final set of samples.

\subsection{Langevin Sampling}\label{appendix:Langevin_sampling}

The gradient of the marginal energy \wrt visible units is,
\begin{align}
    \frac{\partial \tilde{E}(\mathbf{v})}{\partial \mathbf{v}} = \frac{\mathbf{v} - \bm{\mu} }{ \bm{\sigma}^2 } - \frac{W\, \text{Sigmoid} \left( W^{\top} \frac{\mathbf{v}}{\bm{\sigma}^2} + \mathbf{b} \right)}{\bm{\sigma}^2}.
\end{align}

The cosine scheduler for annealing the step size is,
\begin{align}\label{eq:cosine_scheduler}
    \alpha_{k} &= \text{CosineScheduler}(k, K, \alpha_0) = \frac{1}{2} \alpha_0 \left( 1 + \cos \left( \frac{k}{K} \pi \right) \right) 
\end{align}
where $\alpha_{k}$ is the $k$-th step size, $\alpha_{0}$ is the initial step size, and $K$ is the total number of steps.

The derivation of the Metropolis adjustment for Langevin sampling is as follows,
\begin{align}
    \tilde{A}(\mathbf{v}^{(t)}, \mathbf{v}^{(t-1)}) & = \min \left(1, \frac{ p(\mathbf{v}^{(t)}) q( \mathbf{v}^{(t-1)} \vert \mathbf{v}^{(t)}) }{ p(\mathbf{v}^{(t-1)}) q( \mathbf{v}^{(t)} \vert \mathbf{v}^{(t-1)}) } \right) \nonumber \\
    & = \min \left(1, \frac{\exp\left(-\tilde{E}_{\theta}(\mathbf{v}^{(t)})\right) \exp\left( -\frac{1}{4 \alpha_{t}} \left\Vert \mathbf{v}^{(t-1)} - \mathbf{v}^{(t)} + \alpha_{t} \frac{\partial \tilde{E}(\mathbf{v}^{(t)})}{\partial \mathbf{v}} \right\Vert^2 \right) }{\exp\left(-\tilde{E}_{\theta}(\mathbf{v}^{(t-1)})\right) \exp\left( -\frac{1}{4 \alpha_{t}} \left\Vert \mathbf{v}^{(t)} - \mathbf{v}^{(t-1)} + \alpha_{t} \frac{\partial \tilde{E}(\mathbf{v}^{(t-1)})}{\partial \mathbf{v}} \right\Vert^2 \right) } \right) \nonumber \\
    & = \min \left(1, \frac{\exp\left( -\tilde{E}_{\theta}(\mathbf{v}^{(t)}) -\frac{1}{4 \alpha_{t}} \left\Vert \mathbf{v}^{(t-1)} - \mathbf{v}^{(t)} + \alpha_{t} \frac{\partial \tilde{E}(\mathbf{v}^{(t)})}{\partial \mathbf{v}} \right\Vert^2 \right) }{\exp\left(-\tilde{E}_{\theta}(\mathbf{v}^{(t-1)}) -\frac{1}{4 \alpha_{t}} \left\Vert \mathbf{v}^{(t)} - \mathbf{v}^{(t-1)} + \alpha_{t} \frac{\partial \tilde{E}(\mathbf{v}^{(t-1)})}{\partial \mathbf{v}} \right\Vert^2 \right) } \right).
\end{align}

\subsection{Gibbs-Langevin Sampling}\label{appendix:Gibbs_Langevin_sampling}

We now derive the Metropolis Adjustment for Gibbs-Langevin sampling. 
At time step t-1, the proposal distribution in Alg. \ref{alg:Gibbs_Langevin} is
\begin{align}
q(\mathbf{v}, \mathbf{h} \vert \mathbf{v}^{(t-1)}, \mathbf{h}^{(t-1)}) &= q(\mathbf{h} \vert \mathbf{v}) q(\mathbf{v} \vert \mathbf{v}^{(t-1)}, \mathbf{h}^{(t-1)}),
\end{align}
where
\begin{align}
q(\mathbf{h} \vert \mathbf{v}) &= \text{Sigmoid} \left( W^{\top} \left( \frac{\mathbf{v}}{\bm{\sigma}^2} \right) + \mathbf{b} \right).
\end{align}
Denoting $\mathbf{v}^{(t-1)} = \tilde{\mathbf{v}}^{(0)}$ and $\mathbf{v} = \tilde{\mathbf{v}}^{(K)}$, we have,
\begin{align}
q(\mathbf{v} \vert \mathbf{v}^{(t-1)}, \mathbf{h}^{(t-1)}) &= q(\tilde{\mathbf{v}}^{(K)} \vert \tilde{\mathbf{v}}^{(0)}, \mathbf{h}^{(t-1)}) \nonumber \\
&= \int \cdots \int \left( \prod_{k=1}^{K} q(\tilde{\mathbf{v}}^{(k)} \vert \tilde{\mathbf{v}}^{(k-1)}, \mathbf{h}^{(t-1)}) \right) \mathrm{d}\tilde{\mathbf{v}}^{(1)} \cdots \mathrm{d}\tilde{\mathbf{v}}^{(K-1)},      
\end{align}
where
\begin{align}
q(\mathbf{v} \vert \tilde{\mathbf{v}}^{(k-1)}, \mathbf{h}^{(t-1)}) &= \mathcal{N} \left(\mathbf{v} \middle\vert \tilde{\mathbf{v}}^{(k-1)} - \alpha_{k} \frac{\partial E(\tilde{\mathbf{v}}^{(k-1)}, \mathbf{h}^{(t-1)})}{\partial \mathbf{v}}, 2 \alpha_{k} I \right) \\
\frac{\partial E(\mathbf{v}, \mathbf{h})}{\partial \mathbf{v}} &= \frac{\mathbf{v} - \bm{\mu} - W\mathbf{h}}{\bm{\sigma}^2}.
\end{align}
The key question here is how to derive the analytical form of $q(\tilde{\mathbf{v}}^{(K)} \vert \tilde{\mathbf{v}}^{(0)}, \mathbf{h}^{(t-1)})$.
The most straightforward way is to compute the multiple integral directly.
By fixing all variables except for $\tilde{\mathbf{v}}^{(k)}$ in Eq. (\ref{eq:proposal_Gibbs_Langevin}), we can integrate out $\tilde{\mathbf{v}}^{(k)}$ analytically via the Gaussian integral trick, \ie, $\int_{-\infty}^{\infty} \exp(-ax^2+bx+c) \mathrm{d}x = \sqrt{\frac{\pi}{a}}\exp(\frac{b^2}{4a} + c)$.
Then by applying the same trick recursively, one can ideally integrate out all $\tilde{\mathbf{v}}^{(1)}, \dots, \tilde{\mathbf{v}}^{(K-1)}$ in an analytical manner.
However, this process is quite involved due to the fact that the integral of $\tilde{\mathbf{v}}^{(k)}$ depends on both $\tilde{\mathbf{v}}^{(k+1)}$ and $\tilde{\mathbf{v}}^{(k-1)}$.

We instead resort to the reparameterization trick.
In particular, at the outer loop step $t$, the $k$-th inner loop step of Langevin sampling is as follows,
\begin{align}
    \tilde{\mathbf{v}}^{(k)} &= \tilde{\mathbf{v}}^{(k-1)} - \alpha_{k} \frac{\partial E(\tilde{\mathbf{v}}^{(k-1)}, \mathbf{h}^{(t-1)})}{\partial \mathbf{v}} + \sqrt{2 \alpha_{k}} \bm{\xi}_{k} \nonumber \\
    &= \tilde{\mathbf{v}}^{(k-1)} - \alpha_{k} \frac{\tilde{\mathbf{v}}^{(k-1)} - \bm{\mu} - W \mathbf{h}^{(t-1)}}{\bm{\sigma}^2} + \sqrt{2 \alpha_{k}} \bm{\xi}_{k} \nonumber \\
    &= \left( \bm{1} - \frac{\alpha_{k}}{\bm{\sigma}^2} \right) \tilde{\mathbf{v}}^{(k-1)} + \alpha_{k} \frac{\bm{\mu} + W \mathbf{h}^{(t-1)}}{\bm{\sigma}^2} + \sqrt{2 \alpha_{k}} \bm{\xi}_{k},
\end{align}
where $\forall k \in \{1, \dots, K\}$, $\bm{\xi}_{k} \sim \mathcal{N}(0, I)$.
This discretization of Langevin dynamics gives a sample path of the distribution $q(\tilde{\mathbf{v}}^{(K)} \vert \tilde{\mathbf{v}}^{(0)}, \mathbf{h}^{(t-1)})$.
We now show that this sample path could be reparameterized as a simpler one which gives the desirable analytical form of $q(\tilde{\mathbf{v}}^{(K)} \vert \tilde{\mathbf{v}}^{(0)}, \mathbf{h}^{(t-1)})$.
To simplify the derivation, we introduce $\bm{\beta}_{k} = \prod_{j=k+1}^{K} \left( \bm{1} - \frac{\alpha_{j}}{\bm{\sigma}^2} \right)$, $\forall k \in \{0, \dots, K-1\}$ and $\bm{\beta}_{K} = \bm{1}$.
Therefore, after $K$ steps, we have,
\begin{align}
    \tilde{\mathbf{v}}^{(K)} &= \left( \bm{1} - \frac{\alpha_{K}}{\bm{\sigma}^2} \right) \tilde{\mathbf{v}}^{(K-1)} + \alpha_{K} \frac{\bm{\mu} + W \mathbf{h}^{(t-1)}}{\bm{\sigma}^2} + \sqrt{2 \alpha_{K}} \bm{\xi}_{K} \nonumber \\
    &= \left( \prod_{k=1}^{K} \left( \bm{1} - \frac{\alpha_{k}}{\bm{\sigma}^2} \right) \right) \tilde{\mathbf{v}}^{(0)} + \sum_{k=1}^{K} \left( \prod_{j=k+1}^{K} \left( \bm{1} - \frac{\alpha_{j}}{\bm{\sigma}^2} \right) \right) \left( \alpha_{k} \frac{\bm{\mu} + W \mathbf{h}^{(t-1)}}{\bm{\sigma}^2} + \sqrt{2 \alpha_{k}} \bm{\xi}_{k} \right) \nonumber \\
    &= \bm{\beta}_0 \tilde{\mathbf{v}}^{(0)} 
    + \left( \sum_{k=1}^{K} \bm{\beta}_{k} \alpha_{k} \right) \frac{\bm{\mu} + W \mathbf{h}^{(t-1)}}{\bm{\sigma}^2}
    + \sum_{k=1}^{K} \bm{\beta}_{k} \sqrt{2 \alpha_{k}} \bm{\xi}_{k} 
\end{align}
Here $\{\bm{\xi}_{k} \vert k=1, \dots, K\}$ are independent random variables from the standard Normal distribution.
Since we know that the linear combination of several independent Gaussian random variables leads to another Gaussian random variable, we have
\begin{align}
    q(\tilde{\mathbf{v}}^{(K)} \vert \tilde{\mathbf{v}}^{(0)}, \mathbf{h}^{(t-1)}) = \mathcal{N}\left(\bm{\beta}_0 \tilde{\mathbf{v}}^{(0)} + \left( \sum_{k=1}^{K} \bm{\beta}_{k} \alpha_{k} \right) \frac{\bm{\mu} + W \mathbf{h}^{(t-1)}}{\bm{\sigma}^2}, \sum_{k=1}^{K} 2 \alpha_{k} \bm{\beta}_{k}^2 \right).
\end{align}
We can compute the acceptance probability,
\begin{align}
    & A((\mathbf{v}^{(t)}, \mathbf{h}^{(t)}), (\mathbf{v}^{(t-1)}, \mathbf{h}^{(t-1)})) \nonumber \\
    & = \min \left(1, \frac{ p(\mathbf{v}^{(t)}, \mathbf{h}^{(t)}) q( \mathbf{v}^{(t-1)}, \mathbf{h}^{(t-1)} \vert \mathbf{v}^{(t)}, \mathbf{h}^{(t)}) }{ p(\mathbf{v}^{(t-1)}, \mathbf{h}^{(t-1)}) q( \mathbf{v}^{(t)}, \mathbf{h}^{(t)} \vert \mathbf{v}^{(t-1)}, \mathbf{h}^{(t-1)}) } \right) \nonumber \\
    & = \min \left(1, \frac{ p(\mathbf{v}^{(t)}, \mathbf{h}^{(t)}) q(\mathbf{h}^{(t-1)} \vert \mathbf{v}^{(t-1)}) q(\mathbf{v}^{(t-1)} \vert \mathbf{v}^{(t)}, \mathbf{h}^{(t)}) }{ p(\mathbf{v}^{(t-1)}, \mathbf{h}^{(t-1)}) q(\mathbf{h}^{(t)} \vert \mathbf{v}^{(t)}) q(\mathbf{v}^{(t)} \vert \mathbf{v}^{(t-1)}, \mathbf{h}^{(t-1)}) } \right) \nonumber \\
    & = \min \left(1, 
    \frac{ \exp\left(-E_{\theta}(\mathbf{v}^{(t)}, \mathbf{h}^{(t)}) - \left\Vert \frac{ \mathbf{v}^{(t-1)} - \bm{\beta}_0 \mathbf{v}^{(t)} - \bm{a} (\bm{\mu} + W \mathbf{h}^{(t)} ) }{ \sqrt{2} \tilde{\bm{\sigma}}} \right\Vert^2 \right) q(\mathbf{h}^{(t-1)} \vert \mathbf{v}^{(t-1)}) }
    { \exp\left(-E_{\theta}(\mathbf{v}^{(t-1)}, \mathbf{h}^{(t-1)}) - \left\Vert \frac{ \mathbf{v}^{(t)} - \bm{\beta}_0 \mathbf{v}^{(t-1)} - \bm{a} (\bm{\mu} + W \mathbf{h}^{(t-1)} ) }{ \sqrt{2} \tilde{\bm{\sigma}}} \right\Vert^2 \right) q(\mathbf{h}^{(t)} \vert \mathbf{v}^{(t)}) } 
    \right),
\end{align}
where $q(\mathbf{h}_j = 1 \vert \mathbf{v}) = \left[\text{Sigmoid} \left( W^{\top} \frac{\mathbf{v}}{\bm{\sigma}^2} + \mathbf{b} \right) \right]_{j}$, $\bm{a} = \frac{\sum_{k=1}^{K} \bm{\beta}_{k} \alpha_{k}}{\bm{\sigma}^2}$, and $\tilde{\bm{\sigma}}^2 = \sum_{k=1}^{K} 2 \alpha_{k} \bm{\beta}_{k}^2$.

\subsection{Learning}\label{appendix:learning}

We derive the detailed gradients of the marginalized log likelihood of visible units \wrt model parameters as below.
\begin{align}
    \nabla W_{ij} &= \left\langle \frac{\mathbf{v}_{i}}{\bm{\sigma}_i^2} \left[\text{Sigmoid} \left( W^{\top} \frac{\mathbf{v}}{\sigma^2} + \mathbf{b} \right) \right]_{j} \right\rangle_{d} - \left\langle \frac{\mathbf{v}_{i}}{\bm{\sigma}_i^2} \left[\text{Sigmoid} \left( W^{\top} \frac{\mathbf{v}}{\sigma^2} + \mathbf{b} \right) \right]_{j} \right\rangle_{m} \label{eq:gradients_marginal_GRBM_1} \\
    \nabla \mu_{i} &= \left\langle \frac{\mathbf{v}_{i} - \mu_{i}}{\bm{\sigma}_i^2} \right\rangle_{d} - \left\langle \frac{\mathbf{v}_{i} - \mu_{i}}{\bm{\sigma}_i^2} \right\rangle_{m} \label{eq:gradients_marginal_GRBM_2} \\
    \nabla \log\bm{\sigma}_{i}^2 &= \left\langle \frac{(\mathbf{v}_{i} - \mu_{i})^2}{2\bm{\sigma}_{i}^2} - \frac{\sum_{j} \mathbf{v}_{i} W_{ij} \left[\text{Sigmoid} \left( W^{\top} \frac{\mathbf{v}}{\sigma^2} + \mathbf{b} \right) \right]_{j}}{\bm{\sigma}_{i}^2} \right\rangle_{d} \nonumber \\
    & \quad - \left\langle \frac{(\mathbf{v}_{i} - \mu_{i})^2}{2\bm{\sigma}_{i}^2} - \frac{\sum_{j} \mathbf{v}_{i} W_{ij} \left[\text{Sigmoid} \left( W^{\top} \frac{\mathbf{v}}{\sigma^2} + \mathbf{b} \right) \right]_{j}}{\bm{\sigma}_{i}^2} \right\rangle_{m} \label{eq:gradients_marginal_GRBM_3} \\
    \nabla \mathbf{b}_{i} &= \langle \left[\text{Sigmoid} \left( W^{\top} \frac{\mathbf{v}}{\sigma^2} + \mathbf{b} \right) \right]_{i} \rangle_{d} - \langle \left[\text{Sigmoid} \left( W^{\top} \frac{\mathbf{v}}{\sigma^2} + \mathbf{b} \right) \right]_{i} \rangle_{m}. \label{eq:gradients_marginal_GRBM_4}
\end{align}

\section{More Experimental Results}
\label{app:more-results}

For all experiments, we set the initial variances of GRBMs to be $1$, clip the gradient norm to be no larger than $10$, and use SGD with neither momentum nor weight decay.
We divide the total energy of a mini-batch by the batch size so that we are minimizing the average negative log likelihood.
We also decay the learning rate of SGD from the initial value $0.01$ to $0$ using the same cosine scheduler as described in Eq. (\ref{eq:cosine_scheduler}).
The burn-in step in CD learning is set to $0$, \ie, we do not discard any samples from any Markov chains.
For all experiments involving images, we standardize the input image by subtracting the pixel-wise mean and dividing by the pixel-wise standard deviation.
For color images, the subtraction and division is performed channel-wise.

\subsection{Gaussian Mixture Densities}\label{appendix:GMM}

The batch size and the hidden size are set to $100$ and $256$ respectively.
We adjust at every step whenever Metropolis adjustmentment is used as the experiments with Gaussian mixture densities are fast.
Although smaller hidden size could work, but we found this size makes learning converges stably for all sampling algorithms.
To ensure a fair comparison, we train all GRBMs for 50K epochs and use the last model to draw the density plots and samples, despite the learning processes with most of inference algorithms converge within 5K to 10K epochs.

\subsection{Ablation Study on MNIST}\label{appendix:ablation}

In this part, we perform ablation study on MNIST to investigate the effect of several important factors.
In all experiments on MNIST, we set batch size to $512$ and the number of epochs to $3000$.
First, we vary the CD step and the hidden size while fixing the other hyperparameters.
The results are shown in Table \ref{tbl:MNIST_ablation_1}.
We found that $4096$ hidden size and $100$ CD steps work the best on MNIST.
More CD steps would potentially be better but take longer time to train.
Then we turn to study the number of Langevin sampling steps, the adjust step size, the initial Langevin step size, and its annealing. 
Here annealing means we decay the initial Langevin step size to $0$ following the cosine scheduler as training goes on.
The results are shown in Table \ref{tbl:MNIST_ablation_2}.
We can see that the larger the initial Langevin step size, the better the performance.
But values larger than $0.04$ would sometimes make the sampling numerically fail. 
The more the Langevin steps, the better the performance would be.
Again, it comes with more computational cost with more Langevin steps.
We also find that it may not be necessary to adjust at every step and annealing the initial step size slightly improves the performance of Gbbis-Langevin with adjustment.

\subsection{More Visual Results}\label{appendix:more_vis}

We train 3K epochs for experiments on both FashionMNIST and CelebA-32 datasets.
For CeleA-2K-64, we train 4K epochs.
The batch size on FashionMNIST and CelebA-32 is $512$ whereas the batch size on CeleA-2K-64 is $100$.

\begin{figure}[t]
\centering
\captionsetup[subfigure]{justification=centering}
\begin{subfigure}{.19\textwidth}
    \centering
    \includegraphics[width=.99\linewidth]{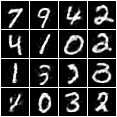}  
    \caption{Gibbs \\~}
\end{subfigure}
\begin{subfigure}{.19\textwidth}
    \centering
    \includegraphics[width=.99\linewidth]{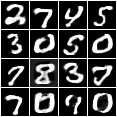}
    \caption{Langevin \\~}
\end{subfigure}
\begin{subfigure}{.19\textwidth}
    \centering
    \includegraphics[width=.99\linewidth]{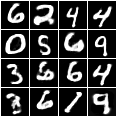}
    \caption{Langevin \\ w. adjust}
\end{subfigure}
\begin{subfigure}{.19\textwidth}
    \centering
    \includegraphics[width=.99\linewidth]{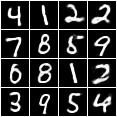}
    \caption{Gibbs-Langevin \\~}
\end{subfigure}
\begin{subfigure}{.19\textwidth}
    \centering
    \includegraphics[width=.99\linewidth]{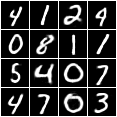}
    \caption{Gibbs-Langevin w. adjust}
\end{subfigure}
\caption{Samples from GRBMs learned with different sampling algorithms on MNIST.}
\label{fig:MNIST_sampling_comparison}
\end{figure}

\begin{figure}[t]
\centering
\begin{subfigure}{.48\textwidth}
    \centering
    \includegraphics[width=.99\linewidth]{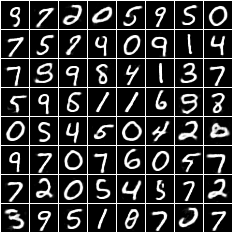}  
    % \caption{}
\end{subfigure}
\begin{subfigure}{.48\textwidth}
    \centering
    \includegraphics[width=.99\linewidth]{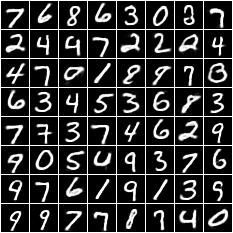}
    % \caption{}
\end{subfigure}
\caption{More samples from the learned GRBM (Gibbs-Langevin) on MNIST.}
\label{fig:MNIST_more}
\end{figure}

\begin{figure}[t]
\centering
\begin{subfigure}{.48\textwidth}
    \centering
    \includegraphics[width=.99\linewidth]{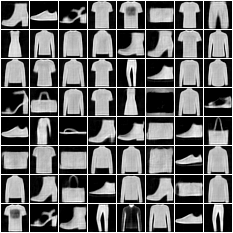}
    % \caption{}
\end{subfigure}
\begin{subfigure}{.48\textwidth}
    \centering
    \includegraphics[width=.99\linewidth]{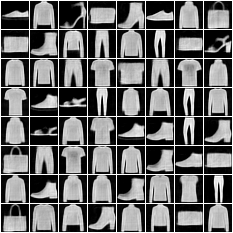}
    % \caption{}
\end{subfigure}
\caption{More samples from the learned GRBM (Gibbs-Langevin) on FashionMNIST.}
\label{fig:FashionMNIST_more}
\end{figure}

\begin{figure}[t]
\centering
\begin{subfigure}{.48\textwidth}
    \centering
    \includegraphics[width=.99\linewidth]{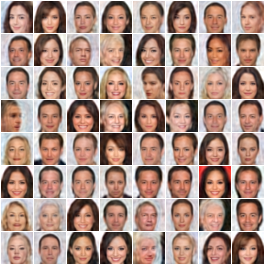}  
    % \caption{}
\end{subfigure}
\begin{subfigure}{.48\textwidth}
    \centering
    \includegraphics[width=.99\linewidth]{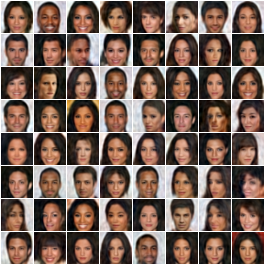}
    % \caption{}
\end{subfigure}
\caption{More samples from the learned GRBM (Gibbs-Langevin) on CelebA-32.}
\label{fig:CelebA_more}
\end{figure}

\begin{figure}[t]
\centering
\begin{subfigure}{.48\textwidth}
    \centering
    \includegraphics[width=.99\linewidth]{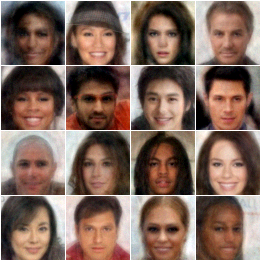}  
    % \caption{}
\end{subfigure}
\begin{subfigure}{.48\textwidth}
    \centering
    \includegraphics[width=.99\linewidth]{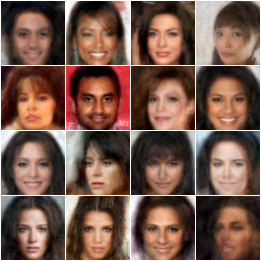}
    % \caption{}
\end{subfigure}
\caption{More samples from the learned GRBM (Gibbs-Langevin) on CelebA-2K-64.}
\label{fig:CelebA2K_more}
\end{figure}

We show the samples drawn from the best GRBMs learned with different sampling methods in Fig. \ref{fig:MNIST_sampling_comparison}.
It is clear that samples corresponding to Gibbs-Langevin have better visual qualities than those from Langevin and Gibbs.
We also show more results of GRBMs learned with Gibbs-Langevin in Fig. \ref{fig:MNIST_more}, Fig. \ref{fig:FashionMNIST_more}, Fig. \ref{fig:CelebA_more}, and Fig. \ref{fig:CelebA2K_more}.

\begin{table}[h]
    \centering
    {%
    \begin{tabular}{@{}l|cc|c@{}}
    \hline
    \toprule
    Methods & CD Step & Hidden Size & FID \\
    \midrule
    \midrule
    Gibbs-Langevin wo. Adjust & 50 & 2048 & 32.33 \\
    Gibbs-Langevin wo. Adjust & 50 & 4096 & 21.02 \\    
    Gibbs-Langevin wo. Adjust & 50 & 8192 & 22.05 \\
    Gibbs-Langevin wo. Adjust & 100 & 4096 & \textbf{17.49} \\
    \bottomrule
    \end{tabular}
    }
    \captionof{table}{Ablation study of the hidden size and the number of CD steps on MNIST dataset.}
    \label{tbl:MNIST_ablation_1}
\end{table}

\begin{table}[h]
    \centering
    {%
    \begin{tabular}{@{}l|cccc|c@{}}
    \hline
    \toprule
    Methods & \begin{tabular}{@{}c@{}} Langevin \\ Step $K$ \end{tabular} & \begin{tabular}{@{}c@{}} Langevin \\Step Size $\alpha_0$\end{tabular} & Anneal $\alpha_0$ & \begin{tabular}{@{}c@{}} Adjust \\ Step $\eta$ \end{tabular} & FID \\
    \midrule
    \midrule
    Gibbs-Langevin wo. Adjust & 1  & 20 & \xmark & - & 35.08 \\
    Gibbs-Langevin wo. Adjust & 5  & 20 & \xmark & - & 19.00 \\
    Gibbs-Langevin wo. Adjust & 10 & 20 & \xmark & - & \textbf{17.49} \\
    Gibbs-Langevin wo. Adjust & 10 & 10 & \xmark & - & 21.05 \\
    Gibbs-Langevin wo. Adjust & 10 & 5  & \xmark & - & 25.67 \\
    Gibbs-Langevin w.  Adjust & 10 & 20 & \xmark & 0 & 21.31 \\
    Gibbs-Langevin w.  Adjust & 10 & 20 & \xmark & 25 & 21.25 \\
    Gibbs-Langevin w.  Adjust & 10 & 20 & \xmark & 50 & 20.64 \\
    Gibbs-Langevin w.  Adjust & 10 & 20 & \cmark & 50 & 19.27 \\
    \bottomrule
    \end{tabular}
    }
    \captionof{table}{Ablation study of the number of Langevin steps $K$, the initial Langevin step size $\alpha_0$, annealing of the initial Langevin step size, and the Metropolis adjust step $\eta$ on MNIST dataset. All runs use 100 CD steps.}
    \label{tbl:MNIST_ablation_2}
\end{table}

\end{document}